\pdfoutput=1

\documentclass[11pt]{article}

\usepackage[preprint]{acl}

\usepackage{times}
\usepackage{latexsym}
\usepackage{booktabs}
\usepackage{soul}
\usepackage{tabularx}

\usepackage{pifont}

\usepackage{xcolor}
\definecolor{purpleface}{HTML}{9CA4F5}
\definecolor{purpleedge}{HTML}{6B76F7}
\definecolor{tealface}{HTML}{56DBBE}
\definecolor{tealedge}{HTML}{1BCCAA}
\usepackage{amsmath}
\usepackage{amssymb}
\usepackage{float}
\usepackage{placeins}
\usepackage{multirow}
\usepackage[T1]{fontenc}

\usepackage[utf8]{inputenc}

\usepackage{microtype}

\usepackage{inconsolata}

\usepackage{graphicx}

\usepackage{amsmath}
\usepackage{subcaption}

\title{INTRYGUE: Induction-Aware Entropy Gating for Reliable RAG Uncertainty Estimation}

\author{
 \textbf{Alexandra Kuleshova\textsuperscript{1}},
 \textbf{Andrei Volodichev\textsuperscript{1}},
 \textbf{Daria Kotova\textsuperscript{1}},
 \textbf{Alexey Zaytsev\textsuperscript{1, 2}},
 \\
 \textsuperscript{1}Applied AI Institute, 
 \textsuperscript{2}Risk AI Research Lab, 
\\
  \small{    \textbf{Correspondence:} \href{mailto:email@domain}{bazarovaai.239@gmail.com}
 }
}

\begin{document}
\maketitle
\begin{abstract}

While retrieval-augmented generation (RAG) enhances LLM performance, it does not eliminate hallucinations, making accurate detection essential. Uncertainty-based methods are attractive for this purpose because they can be integrated into real-world pipelines with little overhead. One of the most widely used uncertainty signals is predictive entropy. We show, however, that entropy can be unreliable in RAG settings and trace this limitation to two opposing internal effects. Induction heads, which copy patterns from earlier context, causally support correct responses and lower predictive entropy, but they also appear to co-activate entropy neurons that push it back up. As a result, correct, context-grounded responses can still receive high uncertainty scores. To address this, we propose INTRYGUE (Induction-Aware Entropy Gating for Uncertainty Estimation), a training-free, mechanistically grounded method that gates predictive entropy by an attention-based estimate of induction-head activity. Evaluated across four RAG-style benchmarks and six open-source LLMs (4B to 13B parameters), INTRYGUE performs competitively against a wide range of baselines, matching or exceeding the strongest of them in most settings. Our findings suggest that hallucination detection in RAG benefits from combining predictive uncertainty with interpretable internal signals of context utilization\footnote{Our code is available at the following \href{https://github.com/kuleshovaai/intrygue}{link}}. 
\end{abstract}

\section{Introduction}

Large language models (LLMs) are widely deployed~\cite{singh2025openai, minaee2024large}, but their static memory is often insufficient. This is typically addressed with retrieval-augmented generation (RAG), which provides the model with documents retrieved from an external corpus~\cite{fan2024survey, lewis2020retrieval}. Nonetheless, hallucinations persist even when relevant evidence is present in the context~\cite{zhang2025siren}, so reliable detection methods remain essential for high-stakes applications.

\begin{figure}[t!]
    \centering
    \includegraphics[width=\linewidth]{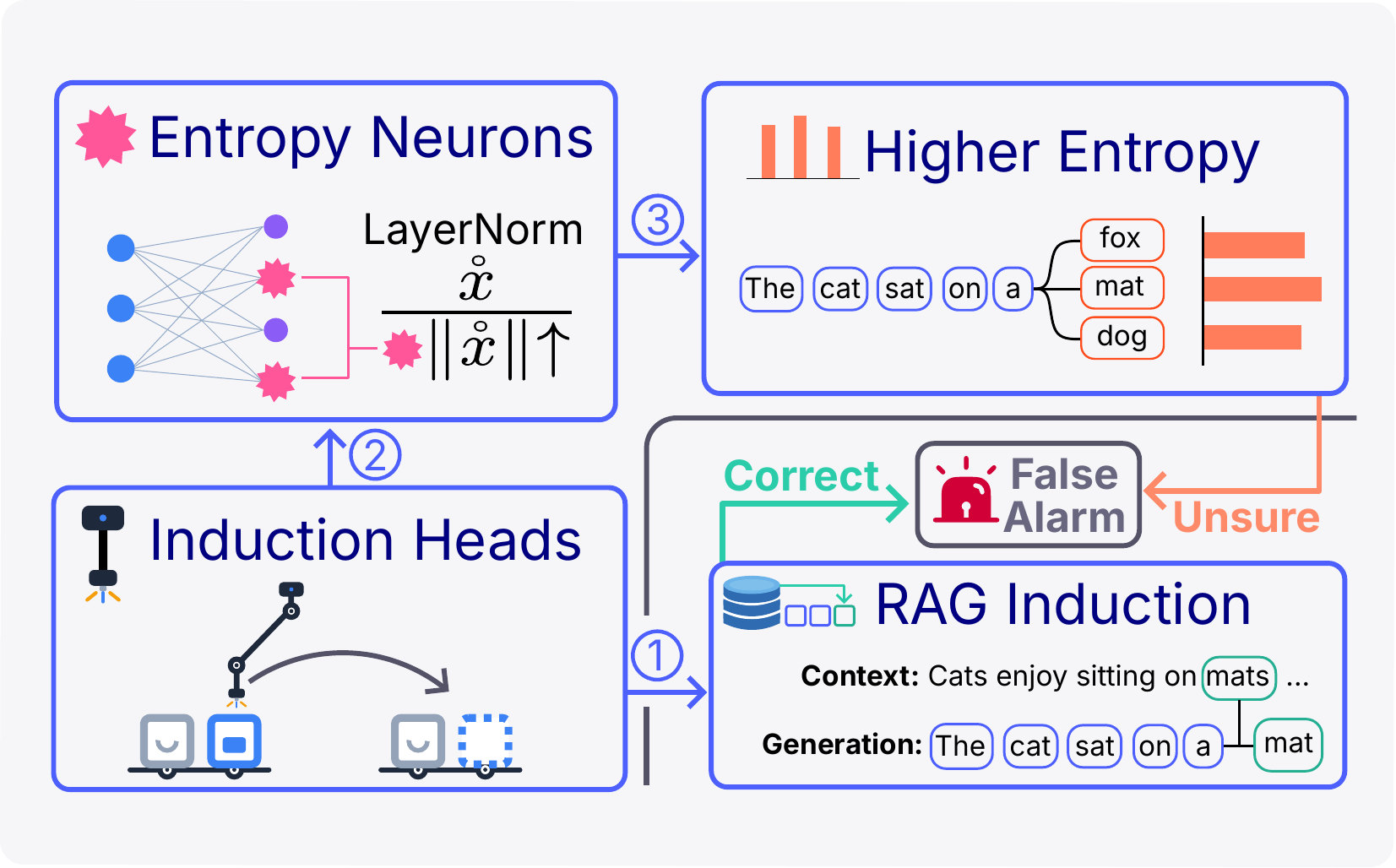} 
    \caption{Opposing effects of induction heads on RAG uncertainty. \ding{192} Induction heads copy patterns from earlier context, supporting the correct answer and lowering predictive entropy. \ding{193} They also co-activate entropy neurons, which inflate the LayerNorm denominator and \ding{194} raise predictive entropy. We argue that this second path can result in high uncertainty scores for context-grounded outputs, leading to false alarms.}
    \label{fig:intro} 
\end{figure}

Uncertainty estimation (UE) provides a natural basis for such detection: rather than relying on external models or multi-step verification~\cite{es2024ragas, friel2023chainpoll}, it derives a confidence score from the model's own predictions~\cite{fomicheva2020unsupervised, malinin2020uncertainty, kuhn2023semantic, qiu2024semantic, qiao2026lowest}. Unfortunately, UE methods show limited effectiveness in RAG, failing to properly reflect the interaction between the retrieved context and the generated response~\cite{soudani2025uncertainty}. A few studies incorporate this interaction into the uncertainty estimates, at the cost of auxiliary models and additional generations~\cite{soudani2025uncertainty, perezuncertainty}. A separate line of work forgoes uncertainty scores altogether, deriving explicit signals of context utilization from the LLM's internal states~\cite{chuang2024lookback} and, in particular, its mechanistic properties~\cite{yeh2026lumina, sunredeep}. These works show that certain mechanisms within LLMs are associated with response faithfulness. However, they use this knowledge to build dedicated hallucination detection features rather than to examine whether these mechanisms interfere with the established confidence measures in RAG.

We address this gap by investigating a mechanistic limitation of predictive entropy~\cite{fomicheva2020unsupervised}~---~a widely used baseline for hallucination detection~\cite{nikitin2024kernel, fadeeva2025faithfulness, qiao2026lowest}~---~in RAG scenarios. We find that correct responses produced in these settings are associated with activation of induction heads, attention heads that copy and continue patterns from earlier in the context~\cite{elhage2021mathematical}. While ablation indicates that these heads support the correct response and lower predictive entropy, they also appear to co-activate entropy neurons~\cite{stolfo2024confidence}, an opposing mechanism that pushes it back up. We argue that these competing effects weaken the link between entropy and correctness, undermining such entropy-based uncertainty estimates in RAG (Figure~\ref{fig:intro}).

To address this issue, we introduce \textbf{\textcolor{purpleedge}{INTRYGUE}} (Induction-Aware Entropy Gating for Uncertainty Estimation), a training-free, mechanistically grounded method that adjusts entropy scores based on the activity of induction heads. Both signals are read from the same forward pass that produces the response, so no auxiliary model and no additional generations are required. INTRYGUE accounts for induction-driven entropy inflation and improves entropy's effectiveness as a hallucination signal in RAG settings.

Our contributions:
\begin{itemize}
\item We show that response correctness in RAG settings is associated with the activation of induction heads, and provide causal evidence that these heads support correct responses and lower predictive entropy;
\item We validate that the connection between induction heads and entropy neurons, previously observed only in controlled induction settings involving exact token repetition and short naturally occurring n-gram repetitions, extends to RAG-based decoding, where induction-head activity co-activates entropy neurons that inflate predictive entropy;
\item Together, our findings indicate that accurate RAG generation can involve two interconnected mechanisms that affect model confidence in opposing directions, an interaction that we argue reduces the reliability of entropy-based uncertainty estimates;
\item Motivated by these findings, we propose \textbf{\textcolor{purpleedge}{INTRYGUE}}, a training-free method that gates predictive entropy by an attention-based estimate of induction-head activity to account for induction-associated entropy inflation;
\item We evaluate \textbf{\textcolor{purpleedge}{INTRYGUE}} against a broad range of uncertainty-estimation and RAG-specific hallucination-detection baselines across four benchmarks and six models ranging from 4B to 13B parameters, observing competitive performance across the evaluated settings.
\end{itemize}

\section{Background}

\subsection{Induction Heads}

Induction heads, which contribute to in-context learning in Transformers~\cite{olsson2022context}, were formalized by~\citet{elhage2021mathematical}. They implement a strict sequence-copying heuristic: given a token pattern $(t_A, t_B, \dots, t_A)$, the head attends strongly to $t_B$ to predict it as the subsequent token. To empirically identify them, one evaluates the model on a randomly generated, repeated sequence $S = (s_1, \dots, s_L, s_1, \dots, s_L)$ and computes an \textit{induction score}. This score is defined as the average attention mass allocated to the correct historical offsets of $S$:
\begin{equation}\label{eq:ind_score}
    \operatorname{InductionScore}(A^{(l, h)}) = \frac{1}{L} \sum_{j = 1}^L \alpha_{L+j, 1+j}^{(l, h)},
\end{equation}
where $A^{(l, h)}$ is the attention matrix of head $h$ at layer $l$, and $\alpha_{t, k}^{(l, h)}$ denotes the attention weight from the current position $t$ to the past position $k$. 

\subsection{Entropy Neurons}

In this paper, the term \textit{neurons} refers to individual entries of MLP hidden states:
\begin{equation}
    \mathrm{MLP}(\mathbf{x}) = \mathbf{W}_{\mathrm{out}}\sigma(\mathbf{W}_{\mathrm{in}}\mathbf{x} + \boldsymbol{\beta}_{\mathrm{in}}) + \boldsymbol{\beta}_{\mathrm{out}},
\end{equation}
where $\mathbf{x} \in \mathbb{R}^{d_{\text{model}}}$ is a normalized
residual stream hidden state, $\mathbf{W}_{\mathrm{out}}^\top, \mathbf{W}_{\mathrm{in}} \in \mathbb{R}^{d_{\mathrm{mlp}} \times d_{\mathrm{model}}}$ are learned weight matrices, $\boldsymbol{\beta}_{\mathrm{in}} \in  \mathbb{R}^{d_{\mathrm{mlp}}}, \, \boldsymbol{\beta}_{\mathrm{out}} \in \mathbb{R}^{d_{\mathrm{model}}}$ are learned bias vectors, and $\sigma$ is an activation function. In this work, we consider only the MLP of the final transformer block. The output of this block is then passed through layer normalization\footnote{While modern LLMs utilize RMSNorm (which omits mean-centering), the fundamental mechanism remains the same: the normalization denominator scales with the magnitude of the activations.}:
\begin{equation}\label{eq:layernorm}
    \mathrm{LN}(\mathbf{x}) = \frac{\mathbf{x} - m(\mathbf{x})}{\sqrt{\mathrm{Var}(\mathbf{x}) + \epsilon}} \odot \boldsymbol{\gamma} + \boldsymbol{\beta},
\end{equation}
where $m(\mathbf{x})$ and $\mathrm{Var}(\mathbf{x})$ denote the mean and variance of the entries of $\mathbf{x}$, and $\boldsymbol{\gamma}, \boldsymbol{\beta} \in \mathbb{R}^{d_{\mathrm{model}}}$ are learned scale and bias parameters. The resulting vector is finally transformed into the logits by the unembedding matrix $\mathbf{W}_{\mathrm{U}} \in \mathbb{R}^{|\mathcal{V}| \times d_{\mathrm{model}}}$, with $|\mathcal{V}|$ being the vocabulary size, followed by a softmax to obtain a probability distribution over $\mathcal{V}$.

Entropy neurons~\cite{gurnee2024universal} are characterized by a large norm of their corresponding column $\mathbf{w}_{\mathrm{out}}$ of $\mathbf{W}_{\mathrm{out}}$ and a near-uniform shift to the logits, quantified by the $\operatorname{LogitVar}$ score:
\begin{equation}\label{eq:logitvar}
    \operatorname{LogitVar}(\mathbf{w}_{\mathrm{out}}) = \operatorname{Var}\left(\frac{\mathbf{W}_{\mathrm{U}}\mathbf{w}_{\mathrm{out}}}{\|\mathbf{W}_{\mathrm{U}}\|_{\dim=1}\|\mathbf{w}_{\mathrm{out}}\|}\right),
\end{equation}
where $\| \cdot \|_{\dim=1}$ indicates column-wise norm. The large norm, in turn, affects the LayerNorm scaling, flattening the logit distribution and reducing model confidence, analogously to raising the softmax temperature.

\citet{stolfo2024confidence} suggest that induction heads activate entropy neurons, but their validation was limited to controlled induction settings with synthetic exact token repetition and short naturally occurring n-gram repetitions of fixed length, leaving open whether this mechanism generalizes to broader contexts where the model must reconcile retrieved external knowledge with its own parametric memory.

\section{Empirical Study: Predictive Entropy Underperforms in RAG}\label{empirical_study}
In this section, we investigate why predictive entropy underperforms in RAG by examining the dual role that induction heads play in such scenarios.

\subsection{Predictive Entropy as an Uncertainty Estimate}

\begin{figure}[t!]
    \centering
    \includegraphics[width=\linewidth]{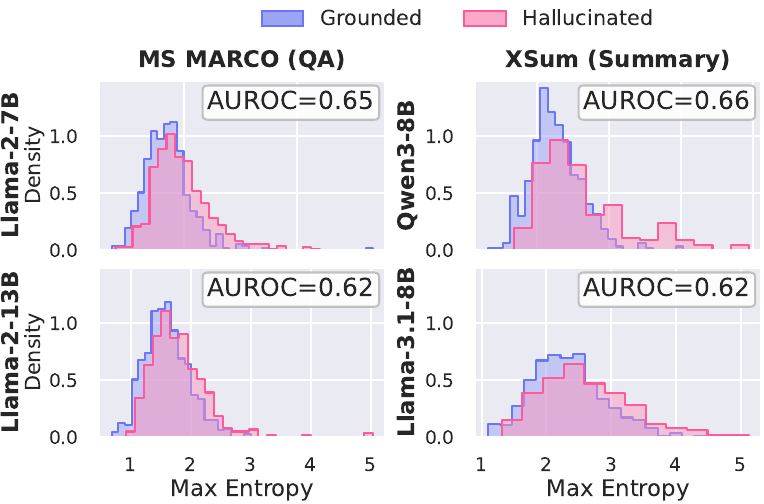} 
    \caption{Maximum Entropy score distributions for hallucinated (pink) and grounded (blue) samples heavily overlap, demonstrating that Maximum Entropy fails to reliably separate hallucinated from grounded LLM responses.}
    \label{fig:entropy_distr} 
\end{figure}

Although predictive entropy is a widely adopted baseline in hallucination detection~\cite{nikitin2024kernel, fadeeva2025faithfulness, qiao2026lowest}, its effectiveness in RAG settings is limited. To illustrate this, we evaluate Maximum Entropy~\cite{fomicheva2020unsupervised}, defined as the maximum token-level predictive entropy over the generated response, on the MS MARCO subset of the RAGTruth dataset~\cite{wu2023ragtruth} and XSum~\cite{narayan2018don}. The results are shown in Figure~\ref{fig:entropy_distr}: the score distributions for hallucinated and non-hallucinated responses overlap heavily, indicating that Maximum Entropy provides only a weak signal of generation errors.

\subsection{Dual Role of Induction Heads}

\begin{figure*}[t!]
    \centering
    \includegraphics[width=\linewidth]{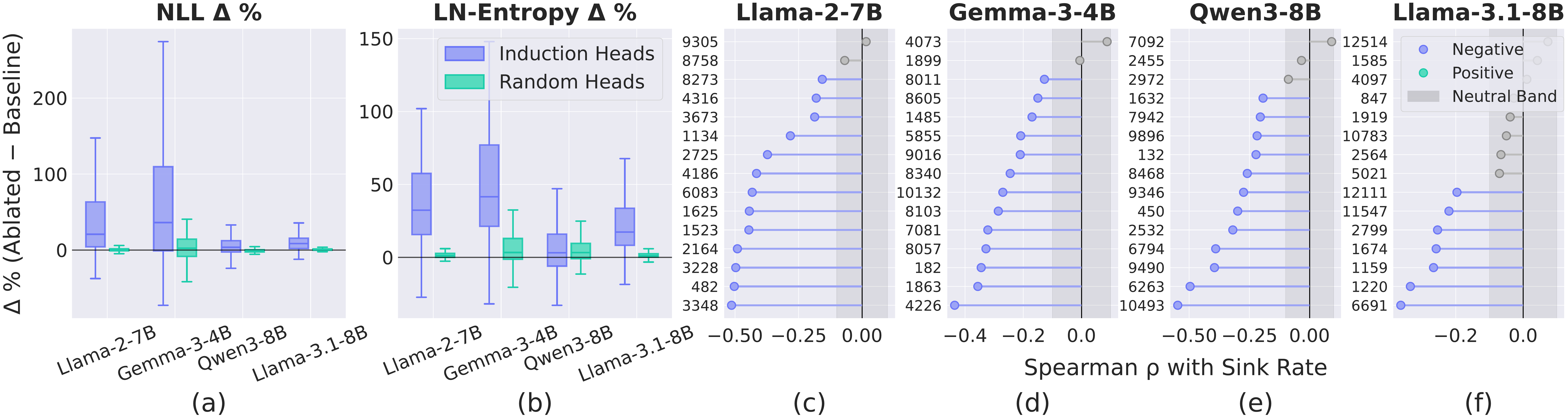} 
    \caption{(a)--(b) Percentage change in Negative Log-Likelihood (NLL) and Entropy following mean ablation of induction heads (blue) versus random heads (green). Ablating induction heads results in significantly higher uncertainty and loss across all models. (c)--(f) Spearman rank correlation ($\rho$) between the SinkRate of induction heads and the activations of the top-15 neurons with the lowest LogitVar scores; this indicates that induction head behavior is associated with the activity of these neurons.
}
    \label{fig:induction_heads_causal_nll}
\end{figure*}

\subsubsection{Induction Heads Pull Confidence Up...}\label{sec:pull_confidence_up}

\paragraph{Correlation.} To investigate whether induction head activation is predictive of hallucination in RAG-based responses, we compute the average SinkRate for the top-5 induction heads\footnote{We compute the InductionScore~\eqref{eq:ind_score} for all heads and select the 5 heads with the greatest scores.} and evaluate its ability to discriminate between hallucinated and grounded responses using the AUROC score. For an attention matrix $A \in \mathbb{R}^{N\times N}$, where $N$ is the total sequence length, the SinkRate\footnote{This definition is motivated by our empirical observation that attention frequently sinks to utility tokens (e.g., punctuation marks and brackets) that are not necessarily located at the beginning of the prompt (see Appendix~\ref{appendix:additional_results}).} is defined~as:
\begin{equation}\label{eq:sinkrate}
    \operatorname{SinkRate}(A, n) = \max_{j} \frac{\sum_{i = N - n + 1}^{N} \alpha_{i, j}}{w_j},
\end{equation}
where $\alpha_{i, j}$ denotes the entries of $A$, $n$ is the response length, and $w_j = \min(n, N - j + 1)$ is a normalization factor that accounts for the number of response tokens able to attend to position $j$. A greater SinkRate indicates deactivation of the induction head, suggesting that induction has not occurred~\cite{elhage2021mathematical}. 

\begin{figure}[t!]
    \centering
    \includegraphics[width=\linewidth]{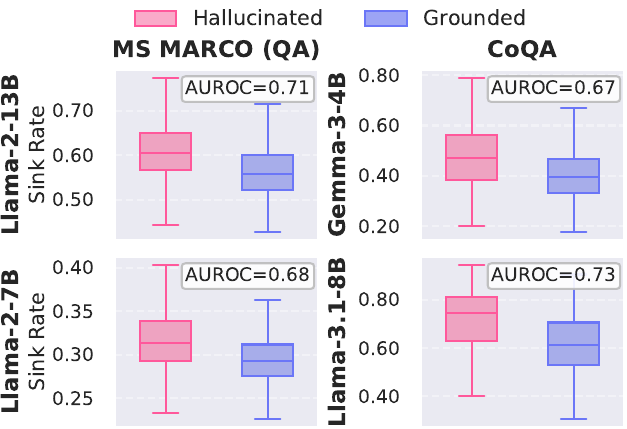} 
    \caption{SinkRate score distributions: hallucinated (pink) vs grounded (blue). AUROC scores and p-values of the Mann-Whitney U-tests are provided in the legend. }
    \label{fig:sink_rate_vs_hallu_label}
\end{figure}

As shown in Figure~\ref{fig:sink_rate_vs_hallu_label}, hallucinated samples consistently exhibit higher SinkRate across all four settings, with statistically significant differences ($p < 10^{-4}$, Mann-Whitney U-test) and AUROC scores ranging from 0.67 to 0.73, implying that the deactivation of induction heads may serve as a potential signal of hallucination in RAG settings.

\paragraph{Causality.} To test whether induction heads causally support correct answer generation, we ran a mean ablation experiment on the CoQA dataset~\citep{reddy2019coqa}. Mean ablation replaces a head's activation with its mean computed over a reference dataset, effectively removing the head's contribution to the forward pass. For correctly answered examples with the greatest induction head activation (SinkRate below the 0.33 quantile), we ablated the top-5 induction heads and compared the effect to ablating 5 randomly chosen heads. 

Figure~\ref{fig:induction_heads_causal_nll}a)-b) shows that ablating induction heads increases loss and entropy relative to randomly chosen heads. This asymmetry indicates that, on correct high-induction responses, these heads causally support the answer: suppressing them both lowers the probability of the correct response and raises predictive entropy.

\subsubsection{...and Push It Back Down}\label{sec:and_push_back_down}
\paragraph{Correlation.} To validate the connection between entropy neurons and induction heads in RAG settings, we examine the correlations between the average SinkRate (over the top-5 induction heads) and the activation scores of the top-15 entropy neurons\footnote{We compute the LogitVar~\eqref{eq:logitvar} for all neurons in the MLP layer of the last transformer block and select the 15 neurons with the lowest scores.}, where the activation score refers to the maximum activation value of a neuron taken across the tokens of the generated response. 

Figure~\ref{fig:induction_heads_causal_nll}c)-f) shows that for observed models, most entropy neurons have a negative Spearman correlation with the SinkRate. These negative correlations indicate that stronger induction-head activity accompanies higher entropy neuron activation, suggesting that the effect reported by \citet{stolfo2024confidence} in controlled setups also appears in RAG-based decoding.

\paragraph{Causality.} 

To establish the causal influence of induction heads on entropy neurons, we performed mean ablation of the top-5 induction heads and measured the resulting change in entropy neuron activations via the $\ell_2$-norm~---~a natural choice given that entropy neurons act by scaling the LayerNorm denominator~\eqref{eq:layernorm}. To focus on samples with strong induction-head activity, we restricted this analysis to those with SinkRate below the 0.33 quantile. As a control, we repeated the same procedure for 5 randomly selected heads.

\begin{figure}[t!]
    \centering
    \includegraphics[width=\linewidth]{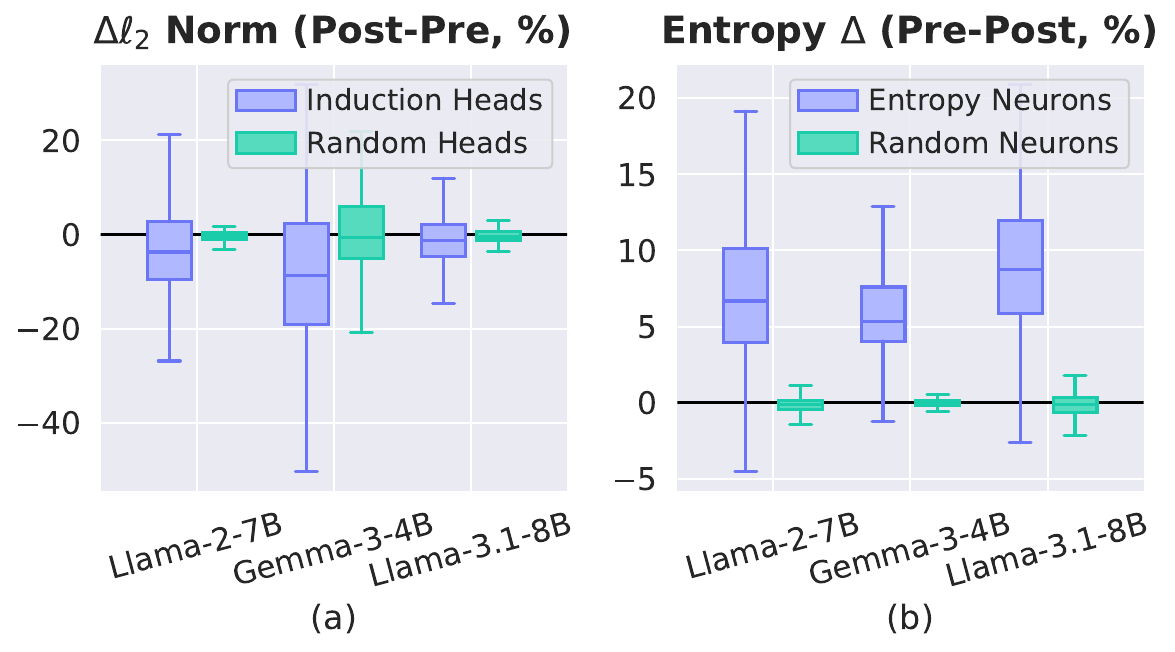}
    \caption{Effects of targeted and random ablations across three models. (a) Relative change in the $\ell_2$-norm of entropy-neuron activations after ablating induction heads or random heads ($\text{Post}-\text{Pre}$, \%); negative values indicate reduced activation after ablation. (b) Relative change in predictive entropy after ablating entropy neurons or random neurons ($\text{Pre}-\text{Post}$, \%); positive values indicate reduced entropy after ablation and thus that the targeted neurons suppress model confidence when active.}
    \label{fig:induction_heads_vs_neurons}
\end{figure}

Figure~\ref{fig:induction_heads_vs_neurons}a) shows that ablating induction heads shifts the distribution of $\ell_2$-norm changes in entropy neuron activations downward, with a negative median in all three models, whereas the median remains near zero for random heads. This indicates that induction-head activity contributes to entropy neuron excitation.

\paragraph{Impact on entropy scores.} We further investigated whether identified entropy neurons causally influence model entropy-based uncertainty by performing targeted ablation experiments on the CoQA dataset. We compared the effect of mean-ablating the top-30 entropy neurons with the ablation of 50 randomly selected neurons within the same MLP layer. Figure~\ref{fig:induction_heads_vs_neurons}b) illustrates that ablating entropy neurons yields a consistent decrease in generation entropy, while random ablation produces near-zero change. These findings confirm that these specific neurons actively modulate the model's predictive confidence.

\FloatBarrier
\begin{table*}[h!]
    \centering
    \caption{AUROC scores for hallucination detection on subsamples of RAGTruth for QA (MS MARCO) and Summarization (CNN/DM). Bold and underline indicate the best and second-best results, respectively. All results are averaged over 5 runs.}
    {\small \setlength{\tabcolsep}{2pt}\renewcommand{\arraystretch}{0.75}
    \begin{tabular}{l|ccc|ccc}\toprule
                       & \multicolumn{3}{c}{MS MARCO}                                                                               & \multicolumn{3}{c}{CNN/DM} \\                                                                    
                      & Mistral-7B                          & Llama-2-7B                          & Llama-2-13B                        & Mistral-7B                          & Llama-2-7B                          & Llama-2-13B                         \\
    \midrule
    $\text{INTRYGUE}_{\text{min-max}}$ & \textbf{0.77 \scriptsize{$\pm$ 0.03}} & \textbf{0.72 \scriptsize{$\pm$ 0.04}} & \underline{0.73 \scriptsize{$\pm$ 0.04}} & \textbf{0.69 \scriptsize{$\pm$ 0.03}} & \underline{0.60 \scriptsize{$\pm$ 0.03}}          & \textbf{0.60 \scriptsize{$\pm$ 0.08}} \\
    $\text{INTRYGUE}_{\text{mean}}$   & 0.71 \scriptsize{$\pm$ 0.06}          & 0.67 \scriptsize{$\pm$ 0.02}          & 0.67 \scriptsize{$\pm$ 0.04}          & \underline{0.65 \scriptsize{$\pm$ 0.03}}          & \textbf{0.62 \scriptsize{$\pm$ 0.08}} & \underline{0.58 \scriptsize{$\pm$ 0.03}}          \\
    \midrule
    Attention Score          & 0.49 \scriptsize{$\pm$ 0.05}          & 0.47 \scriptsize{$\pm$ 0.07}          & 0.51 \scriptsize{$\pm$ 0.05}          & 0.54 \scriptsize{$\pm$ 0.04}          & 0.50 \scriptsize{$\pm$ 0.06}          & 0.54 \scriptsize{$\pm$ 0.03}          \\
    Focus              & 0.67 \scriptsize{$\pm$ 0.06}          & 0.63 \scriptsize{$\pm$ 0.03}          & 0.61 \scriptsize{$\pm$ 0.03}          & 0.62 \scriptsize{$\pm$ 0.04}          & 0.51 \scriptsize{$\pm$ 0.09}          & 0.50 \scriptsize{$\pm$ 0.06}          \\
    LN-Entropy         & 0.65 \scriptsize{$\pm$ 0.06}          & 0.63 \scriptsize{$\pm$ 0.03}          & 0.62 \scriptsize{$\pm$ 0.04}          & 0.64 \scriptsize{$\pm$ 0.04}         & \textbf{0.62 \scriptsize{$\pm$ 0.09}} & 0.56 \scriptsize{$\pm$ 0.05}          \\
    Lookback Lens &	0.73 \scriptsize{$\pm$ 0.03} &	\underline{0.71 \scriptsize{$\pm$ 0.02}}	& \textbf{0.76 \scriptsize{$\pm$ 0.04}} & 0.60 \scriptsize{$\pm$ 0.07} & 0.55 \scriptsize{$\pm$ 0.04} & 	0.49 \scriptsize{$\pm$ 0.01} \\
    LSC            & 0.68 \scriptsize{$\pm$ 0.04}          & 0.57 \scriptsize{$\pm$ 0.03}          & 0.57 \scriptsize{$\pm$ 0.05}          & 0.58 \scriptsize{$\pm$ 0.05}          & 0.49 \scriptsize{$\pm$ 0.07}          & 0.52 \scriptsize{$\pm$ 0.07}          \\ 
    LUMINA   &	0.66 \scriptsize{$\pm$ 0.05}	& \underline{0.71 \scriptsize{$\pm$ 0.04}}	& 0.70 \scriptsize{$\pm$ 0.04} &   0.64 \scriptsize{$\pm$ 0.07}	& 0.54 \scriptsize{$\pm$ 0.06} &	0.54 \scriptsize{$\pm$ 0.04} \\ 
    MaxEntropy         & 0.67 \scriptsize{$\pm$ 0.03}          & 0.65 \scriptsize{$\pm$ 0.06}          & 0.62 \scriptsize{$\pm$ 0.05}          & 0.61 \scriptsize{$\pm$ 0.04}          & \underline{0.60 \scriptsize{$\pm$ 0.04}}          & 0.54 \scriptsize{$\pm$ 0.09}          \\
    MaxProb            & \underline{0.75 \scriptsize{$\pm$ 0.04}}          & 0.69 \scriptsize{$\pm$ 0.03}        & 0.67 \scriptsize{$\pm$ 0.04}          & 0.61 \scriptsize{$\pm$ 0.07}          & 0.51 \scriptsize{$\pm$ 0.07}          & 0.53 \scriptsize{$\pm$ 0.08}          \\
    Perplexity         & 0.64 \scriptsize{$\pm$ 0.05}          & 0.48 \scriptsize{$\pm$ 0.03}          & 0.46 \scriptsize{$\pm$ 0.03}          & 0.49 \scriptsize{$\pm$ 0.08}          & 0.52 \scriptsize{$\pm$ 0.04}          & 0.48 \scriptsize{$\pm$ 0.04}          \\
    RAUQ   & 0.66 \scriptsize{$\pm$ 0.04} &	0.60 \scriptsize{$\pm$ 0.02} &	0.59 \scriptsize{$\pm$ 0.04}  & 0.58 \scriptsize{$\pm$ 0.06}	& 0.50 \scriptsize{$\pm$ 0.10} &	0.51 \scriptsize{$\pm$ 0.05}     \\
    ReDeEP             & 0.72 \scriptsize{$\pm$ 0.03}          & 0.63 \scriptsize{$\pm$ 0.05}          & 0.66 \scriptsize{$\pm$ 0.03}          & 0.63 \scriptsize{$\pm$ 0.08}          & 0.59 \scriptsize{$\pm$ 0.03}          & 0.57 \scriptsize{$\pm$ 0.08}          \\
    \midrule
    EigenScore             & 0.60 \scriptsize{$\pm$ 0.04}          & 0.58 \scriptsize{$\pm$ 0.03}          & 0.58 \scriptsize{$\pm$ 0.04}          & 0.57 \scriptsize{$\pm$ 0.07}          & 0.52 \scriptsize{$\pm$ 0.03}          & 0.51 \scriptsize{$\pm$ 0.06}          \\
    SAR                & 0.67 \scriptsize{$\pm$ 0.04}          & 0.55 \scriptsize{$\pm$ 0.02}          & 0.48 \scriptsize{$\pm$ 0.03}          & 0.57 \scriptsize{$\pm$ 0.04}          & 0.48 \scriptsize{$\pm$ 0.10}          & 0.52 \scriptsize{$\pm$ 0.05}          \\
    Semantic Density & 0.55 \scriptsize{$\pm$ 0.02} & 0.55 \scriptsize{$\pm$ 0.05} & 0.53 \scriptsize{$\pm$ 0.02} & 0.56 \scriptsize{$\pm$ 0.08} & 0.52 \scriptsize{$\pm$ 0.04} & 0.57 \scriptsize{$\pm$ 0.07} \\
    Semantic Entropy & 0.60 \scriptsize{$\pm$ 0.05}          & 0.53 \scriptsize{$\pm$ 0.04}          & 0.62 \scriptsize{$\pm$ 0.04}          & 0.56 \scriptsize{$\pm$ 0.06}          & 0.57 \scriptsize{$\pm$ 0.06}          & 0.52 \scriptsize{$\pm$ 0.03}          \\
    \bottomrule
    \end{tabular}}
    \label{tab:ragtruth}
    \end{table*}

\section{Mechanistic Uncertainty Estimation}

Our findings indicate that standard entropy-based estimators do not account for the opposing effects of induction-head activity, a mechanism associated with faithful RAG generation. To address this, we propose \textbf{\textcolor{purpleedge}{INTRYGUE}} (Induction-Aware Entropy Gating for Uncertainty Estimation). Our method uses aggregated SinkRate~\eqref{eq:sinkrate}, which quantifies activity of induction heads, as a gating term; it down-weights the raw entropy score when induction heads are active, reducing false alarms on grounded responses.

Let $P$ denote the input prompt and let $R = (r_1, \dots, r_n)$ be the generated response sequence of tokens. At each generation step $t$, the model outputs a probability distribution over the vocabulary $\mathcal{V}$. We derive the sequence $\mathcal{E} = \{e_t\}_{t=1}^n$ by calculating the entropy of these distributions to quantify the uncertainty of each predicted token:
\begin{equation}
e_t = -\sum_{v \in \mathcal{V}} \mathbb{P}(v \mid P, R_{<t}) \log \mathbb{P}(v \mid P, R_{<t}).
\end{equation}

Simultaneously, we extract a set of SinkRates $\mathcal{S} = \{s_i\}_{i=1}^k$ defined in Eq.~\ref{eq:sinkrate} corresponding to the top-$k$ induction heads with greatest InductionScore~\eqref{eq:ind_score} values. These rates serve as a mechanistic proxy for quantifying the extent of induction during the generation process.
The hyperparameter $k$ is selected on a validation set.

Let $f$ and $g$ be aggregation functions (e.g., minimum, mean, or maximum) that map $\mathcal{S}$ and $\mathcal{E}$, respectively, to scalars. We employ the aggregated SinkRate as a coefficient that gates the predictive entropy: when induction fails (high SinkRate), the uncertainty signal is left largely intact; when induction is successful (low SinkRate), the signal is down-weighted. We thus formulate our score, INTRYGUE, as:
\begin{equation}\label{eq:intrigue}
    \operatorname{INTRYGUE}(P, R) = \underbrace{ f \big( \{s_i\}_{i=1}^k \big) }_{\substack{\text{Induction} \\ \text{Gate}}} \cdot \underbrace{ g \big( \{e_t\}_{t=1}^n \big) }_{\substack{\text{Predictive} \\ \text{Uncertainty}}}.
 \end{equation} 
In practice, the choice of aggregation functions $f$ and $g$ allows the metric to be tuned to capture either the average uncertainty or the most severe failure points across the generated response.

\section{Results}
\subsection{Experiment Setting}

\paragraph{Datasets and models.} We evaluate on four question answering (QA) and summarization tasks: long-form QA (MS MARCO) and summarization (CNN/DM) from RAGTruth~\cite{wu2023ragtruth}, conversational QA (CoQA~\cite{reddy2019coqa}), and extreme summarization (XSum~\cite{narayan2018don}). Additional details are in Appendix~\ref{sec:appendix_datasets}.

We employ six popular open-source LLMs that provide access to their internal states: gemma-3-4b-it~\citep{gemma_2025}, LLaMA-2-7B-chat, LLaMA-2-13B-chat~\citep{touvron2023llama}, Mistral-7B-Instruct-v0.1~\citep{jiang2023mistral7b}, LLaMA-3.1-8B-Instruct~\citep{grattafiori2024llama}, and Qwen3-8B~\citep{qwen3technicalreport}. As annotated responses for gemma-3-4b-it, LLaMA-3.1-8B, and Qwen3-8B are not available in the RAGTruth dataset, we limit the evaluation of these models to CoQA and XSum.

\paragraph{Baselines.} We compare our method against a comprehensive set of baselines. As direct baselines on predictive entropy, we evaluate LN-Entropy~\citep{malinin2020uncertainty} and MaxEntropy~\citep{fadeeva2024fact}, which aggregate token-level predictive entropy over the generated response using the mean and maximum, respectively. The remaining information-based baselines are AttentionScore~\citep{sriramanan2024llm}, Focus~\citep{zhang2023enhancing}, Lowest Span Confidence (LSC)~\citep{qiao2026lowest}, MaxProb~\citep{fadeeva2024fact}, Perplexity~\citep{fomicheva2020unsupervised}, and RAUQ~\citep{vazhentsev2025uncertainty}. Sampling diversity-based methods include EigenScore~\citep{chen2024inside}, SAR~\citep{duan2024shifting}, Semantic Density~\citep{qiu2024semantic}, and Semantic Entropy~\citep{kuhn2023semantic}. As for RAG-specific methods, we evaluate against ReDeEP~\citep{sunredeep}, LUMINA~\citep{yeh2026lumina}, and Lookback Lens~\citep{chuang2024lookback}. To ensure a fair comparison, we train Lookback Lens on the same data used for hyperparameter selection of the unsupervised methods. Full implementation details are available in Appendix~\ref{app:implementation_details}.

\subsection{Main Results}
\label{sec:main_results}

\begin{table*}[t!]
\centering
\caption{Hallucination detection results (AUROC score) on CoQA and XSum datasets. Bold and underline indicate the best and second-best results, respectively. All results are averaged over 5 runs.}
{\small \setlength{\tabcolsep}{4pt}
\renewcommand{\arraystretch}{0.8}
\begin{tabular}{l|cccccc}\toprule
                   & \multicolumn{6}{c}{CoQA}                                                                                                                                                                                                                                                                                              \\
                   \midrule
                   & Mistral-7B                                        & Llama-2-7B                                        & Llama-2-13B                                       & Llama-3.1-8B                                      & Qwen3-8B                                          & gemma-3-4b                                     \\
        \midrule
$\text{INTRYGUE}_\text{min-max}$    & 0.79 \scriptsize{$\pm$ 0.04}                        & \underline{0.76 \scriptsize{$\pm$ 0.04}}                  & 0.72 \scriptsize{$\pm$ 0.05}                        & 0.78 \scriptsize{$\pm$ 0.08}                        & 0.77 \scriptsize{$\pm$ 0.07}                        & 0.73 \scriptsize{$\pm$ 0.05}                        \\
$\text{INTRYGUE}_{\text{mean}}$       & \underline{0.82 \scriptsize{$\pm$ 0.04}}                 & \textbf{0.79 \scriptsize{$\pm$ 0.03}}               & \textbf{0.81 \scriptsize{$\pm$ 0.03}}               & \underline{0.81 \scriptsize{$\pm$ 0.07}}               & \textbf{0.80 \scriptsize{$\pm$ 0.06}}               & \textbf{0.76 \scriptsize{$\pm$ 0.04}}               \\
\midrule
Attention Score          & 0.55 \scriptsize{$\pm$ 0.03}                        & {\color[HTML]{181D1F} 0.54 \scriptsize{$\pm$ 0.06}} & 0.60 \scriptsize{$\pm$ 0.06}                        & 0.56 \scriptsize{$\pm$ 0.04}                        & {\color[HTML]{181D1F} 0.58 \scriptsize{$\pm$ 0.09}} & {\color[HTML]{181D1F} 0.51 \scriptsize{$\pm$ 0.10}} \\
Focus              & 0.53 \scriptsize{$\pm$ 0.05}                        & 0.33 \scriptsize{$\pm$ 0.02}                        & 0.31 \scriptsize{$\pm$ 0.03}                        & 0.53 \scriptsize{$\pm$ 0.09}                        & 0.42 \scriptsize{$\pm$ 0.04}                        & 0.48 \scriptsize{$\pm$ 0.03}                        \\
LN-Entropy         & \textbf{0.83 \scriptsize{$\pm$ 0.03}}               & 0.74 \scriptsize{$\pm$ 0.03}                        & \underline{0.80 \scriptsize{$\pm$ 0.04}}                  & \underline{0.81 \scriptsize{$\pm$ 0.08}}               & \underline{0.79 \scriptsize{$\pm$ 0.05}}                  & \underline{0.75 \scriptsize{$\pm$ 0.04}}                  \\
Lookback Lens	& 0.64 \scriptsize{$\pm$ 0.03} & 0.74 \scriptsize{$\pm$ 0.03}	& 0.73 \scriptsize{$\pm$} 0.05 &	0.68 \scriptsize{$\pm$ 0.02} &	0.70 \scriptsize{$\pm$ 0.05} &	0.62 \scriptsize{$\pm$ 0.05} \\
LSC            & 0.72 \scriptsize{$\pm$ 0.03}                        & 0.70 \scriptsize{$\pm$  0.03}                       & 0.55 \scriptsize{$\pm$ 0.05}                        & 0.77 \scriptsize{$\pm$ 0.07}                        & 0.75 \scriptsize{$\pm$ 0.06}                        & 0.58 \scriptsize{$\pm$ 0.09}                        \\
LUMINA	& 0.70 \scriptsize{$\pm$ 0.03}	& 0.69 \scriptsize{$\pm$ 0.03} &	0.64 \scriptsize{$\pm$ 0.05} &	0.69 \scriptsize{$\pm$ 0.05} &	0.78 \scriptsize{$\pm$ 0.06}	& 0.70 \scriptsize{$\pm$ 0.04} \\
MaxEntropy         & 0.78 \scriptsize{$\pm$ 0.03}                        & 0.60 \scriptsize{$\pm$ 0.03}                        & 0.65 \scriptsize{$\pm$ 0.06}                        & 0.75 \scriptsize{$\pm$ 0.06}                        & 0.73 \scriptsize{$\pm$ 0.05}                        & 0.70 \scriptsize{$\pm$ 0.07}                        \\
MaxProb            & 0.72 \scriptsize{$\pm$ 0.05}                        & 0.54 \scriptsize{$\pm$  0.03}                       & 0.60 \scriptsize{$\pm$ 0.08}                        & 0.68 \scriptsize{$\pm$ 0.09}                        & 0.50 \scriptsize{$\pm$ 0.09}                        & 0.58 \scriptsize{$\pm$ 0.06}                        \\
Perplexity         & 0.59 \scriptsize{$\pm$ 0.04}                        & 0.52 \scriptsize{$\pm$ 0.05}                        & 0.58 \scriptsize{$\pm$ 0.10}                        & 0.50 \scriptsize{$\pm$ 0.08}                        & 0.43 \scriptsize{$\pm$ 0.07}                        & 0.51 \scriptsize{$\pm$ 0.04}                        \\
RAUQ   & 0.68 \scriptsize{$\pm$ 0.04} & 0.75 \scriptsize{$\pm$ 0.02}	 & 0.75 \scriptsize{$\pm$ 0.06} &	0.69 \scriptsize{$\pm$ 0.08} &	0.62 \scriptsize{$\pm$ 0.07} &	0.58 \scriptsize{$\pm$ 0.05}     \\
ReDeEP             & {0.62 \scriptsize{$\pm$ 0.04}} & {0.67 \scriptsize{$\pm$ 0.06}} & 0.77 \scriptsize{$\pm$ 0.03}                        & 0.67 \scriptsize{$\pm$ 0.05}                        & {\color[HTML]{181D1F} 0.59 \scriptsize{$\pm$ 0.01}} & 0.59 \scriptsize{$\pm$ 0.05}                        \\
\midrule
EigenScore             & {0.69 \scriptsize{$\pm$ 0.05}} & { 0.55 \scriptsize{$\pm$ 0.05}} & {0.60 \scriptsize{$\pm$ 0.03}} & {0.73 \scriptsize{$\pm$ 0.05}} & {0.70 \scriptsize{$\pm$ 0.05}} & 0.71 \scriptsize{$\pm$ 0.05}                        \\
SAR                & 0.72 \scriptsize{$\pm$ 0.02}                        & 0.68 \scriptsize{$\pm$ 0.04}                        & 0.76 \scriptsize{$\pm$ 0.06}                        & 0.64 \scriptsize{$\pm$ 0.06}                        & 0.63 \scriptsize{$\pm$ 0.07}                        & 0.58 \scriptsize{$\pm$ 0.06}                        \\
Semantic Density & 0.61 \scriptsize{$\pm$ 0.03} & 0.59 \scriptsize{$\pm$ 0.05} & 0.63 \scriptsize{$\pm$ 0.04} & 0.59 \scriptsize{$\pm$ 0.08} & 0.63 \scriptsize{$\pm$ 0.06} & 0.58 \scriptsize{$\pm$ 0.08} \\
Semantic Entropy & 0.79 \scriptsize{$\pm$ 0.03}                        & 0.64 \scriptsize{$\pm$ 0.05}                        & 0.72 \scriptsize{$\pm$ 0.05}                        & \textbf{0.82 \scriptsize{$\pm$ 0.06}}                        & 0.72 \scriptsize{$\pm$ 0.05}                        & 0.70 \scriptsize{$\pm$ 0.05}                        \\
\midrule
                   & \multicolumn{6}{c}{XSum}                                                                                                                                                                                                                                                                                              \\
                   \midrule
$\text{INTRYGUE}_{\text{min-max}}$    & \textbf{0.73 \scriptsize{$\pm$ 0.07}}               & \textbf{0.66 \scriptsize{$\pm$ 0.07}}                     & \textbf{0.63 \scriptsize{$\pm$ 0.05}}               & \textbf{0.66 \scriptsize{$\pm$ 0.02}}               & \underline{0.68 \scriptsize{$\pm$ 0.06}}                  & \textbf{0.62 \scriptsize{$\pm$ 0.06}}               \\
$\text{INTRYGUE}_{\text{mean}}$       & \underline{0.70 \scriptsize{$\pm$ 0.03}}                        & 0.63 \scriptsize{$\pm$ 0.03}               & \underline{0.62 \scriptsize{$\pm$ 0.03}   }                     & \underline{0.64 \scriptsize{$\pm$ 0.03} }                       & 0.66 \scriptsize{$\pm$ 0.04}                        & 0.58 \scriptsize{$\pm$ 0.06}                        \\
\midrule
Attention Score        & 0.59 \scriptsize{$\pm$ 0.03}                        & 0.50 \scriptsize{$\pm$ 0.08}                        & 0.56 \scriptsize{$\pm$ 0.04}                        & 0.51 \scriptsize{$\pm$ 0.05}                        & 0.46 \scriptsize{$\pm$ 0.06}                        & 0.50 \scriptsize{$\pm$ 0.02}                        \\
Focus              & 0.63 \scriptsize{$\pm$ 0.08}                        & \underline{0.65 \scriptsize{$\pm$ 0.06}}                        & 0.58 \scriptsize{$\pm$ 0.04}                        & 0.62 \scriptsize{$\pm$ 0.05}                        & 0.61 \scriptsize{$\pm$ 0.03}                        & \underline{0.61 \scriptsize{$\pm$ 0.04}}                        \\
LN-Entropy         & 0.69 \scriptsize{$\pm$ 0.03}                        & 0.61 \scriptsize{$\pm$ 0.02}                        & \underline{0.62 \scriptsize{$\pm$ 0.04} }                       & \underline{0.64 \scriptsize{$\pm$ 0.03}}                       & 0.65 \scriptsize{$\pm$ 0.04}                        & 0.57 \scriptsize{$\pm$ 0.06}                        \\
Lookback Lens &	0.60 \scriptsize{$\pm$ 0.07} &	0.52 \scriptsize{$\pm$ 0.06} &	0.49 \scriptsize{$\pm$ 0.04} &	0.56 \scriptsize{$\pm$ 0.05} &	0.57 \scriptsize{$\pm$ 0.05}	& 0.56 \scriptsize{$\pm$ 0.09} \\
LSC            & 0.64 \scriptsize{$\pm$ 0.05}                        & 0.57 \scriptsize{$\pm$  0.05}                       & 0.58 \scriptsize{$\pm$ 0.04}                        & 0.59 \scriptsize{$\pm$ 0.08}                        & 0.62 \scriptsize{$\pm$ 0.04}                        & 0.53 \scriptsize{$\pm$ 0.07}                        \\
LUMINA & 0.62 \scriptsize{$\pm$ 0.03} & 0.58 \scriptsize{$\pm$ 0.06} &	0.59 \scriptsize{$\pm$ 0.04} &	0.56 \scriptsize{$\pm$ 0.05} &	0.61 \scriptsize{$\pm$ 0.05}	& 0.54 \scriptsize{$\pm$ 0.05} \\
MaxEntropy         & 0.67 \scriptsize{$\pm$ 0.02}                        & 0.60 \scriptsize{$\pm$ 0.06}                        & 0.57 \scriptsize{$\pm$ 0.07}                        & 0.62 \scriptsize{$\pm$ 0.02}                        & 0.67 \scriptsize{$\pm$ 0.06}                        & 0.58 \scriptsize{$\pm$ 0.06}                        \\
MaxProb            & 0.60 \scriptsize{$\pm$ 0.07}                        & 0.58 \scriptsize{$\pm$ 0.05}                        & 0.58 \scriptsize{$\pm$ 0.04}                        & 0.62 \scriptsize{$\pm$ 0.04}                        & 0.63 \scriptsize{$\pm$ 0.04}                        & 0.53 \scriptsize{$\pm$ 0.05}                        \\
Perplexity         & 0.58 \scriptsize{$\pm$ 0.05}                        & 0.56 \scriptsize{$\pm$ 0.05}                        & 0.53 \scriptsize{$\pm$ 0.04}                        & 0.61 \scriptsize{$\pm$ 0.05}                        & 0.63 \scriptsize{$\pm$ 0.03}                        & 0.47 \scriptsize{$\pm$ 0.07}                        \\
RAUQ   & 0.66 \scriptsize{$\pm$ 0.03} &	0.59 \scriptsize{$\pm$ 0.04} &	0.59 \scriptsize{$\pm$ 0.04} &	0.63 \scriptsize{$\pm$ 0.04} &	0.63 \scriptsize{$\pm$ 0.03} &	0.56 \scriptsize{$\pm$ 0.08}        \\
ReDeEP             & 0.62 \scriptsize{$\pm$ 0.04}                        & 0.61 \scriptsize{$\pm$ 0.04}                        & 0.60 \scriptsize{$\pm$ 0.04}                        & 0.63 \scriptsize{$\pm$ 0.02}                        & \textbf{0.69 \scriptsize{$\pm$ 0.05}}               & \textbf{0.62 \scriptsize{$\pm$ 0.06}}               \\
\midrule
EigenScore        & 0.60 \scriptsize{$\pm$ 0.08}                        & 0.60 \scriptsize{$\pm$ 0.05}                        & 0.52 \scriptsize{$\pm$ 0.03}                        & \textbf{0.66 \scriptsize{$\pm$ 0.04}}               & 0.62 \scriptsize{$\pm$ 0.06}                        & \underline{0.61 \scriptsize{$\pm$ 0.08}}                \\
SAR                & 0.66 \scriptsize{$\pm$ 0.03}                        & 0.60 \scriptsize{$\pm$ 0.03}                        & 0.58 \scriptsize{$\pm$ 0.05}                        & 0.63 \scriptsize{$\pm$ 0.04}                        & 0.64 \scriptsize{$\pm$ 0.03}                        & 0.50 \scriptsize{$\pm$ 0.05}                        \\
Semantic Density & 0.63 \scriptsize{$\pm$ 0.03} & 0.63 \scriptsize{$\pm$ 0.06} & 0.59 \scriptsize{$\pm$ 0.05} & 0.55 \scriptsize{$\pm$ 0.06} & 0.60 \scriptsize{$\pm$ 0.07} & 0.58 \scriptsize{$\pm$ 0.06}\\
Semantic Entropy & 0.64 \scriptsize{$\pm$ 0.05}                        & 0.63 \scriptsize{$\pm$ 0.03}                        & \underline{0.62 \scriptsize{$\pm$ 0.02}}                        & 0.61 \scriptsize{$\pm$ 0.05}                        & 0.67 \scriptsize{$\pm$ 0.04}                        & \textbf{0.62 \scriptsize{$\pm$ 0.06}}               \\
\bottomrule
\end{tabular}}\label{tab:coqa_xsum}
\end{table*}

The main experimental results are presented in Tables~\ref{tab:ragtruth}-\ref{tab:coqa_xsum}. We report two implementations of our proposed method: $\operatorname{INTRYGUE}_{\text{min-max}}$ (where $f$ and $g$ are the sequence minimum and maximum, respectively) and $\operatorname{INTRYGUE}_{\text{mean}}$ (where both are the average). 

As shown in Tables~\ref{tab:ragtruth} and \ref{tab:coqa_xsum}, INTRYGUE performs competitively against both general-purpose and RAG-specific baselines, matching or exceeding the strongest of them in most evaluated settings. This suggests that integrating distinct uncertainty signals~---~internal induction dynamics and predictive entropy~---~enables more robust uncertainty quantification.
The optimal INTRYGUE variant depends on response length. For the long, detailed generations in RAGTruth (MS MARCO, CNN/DM) and XSum, averaging over the sequence blurs localized hallucination signals. Consequently, $\operatorname{INTRYGUE}_{\text{min-max}}$ is the stronger of the two variants in these settings, yielding the most pronounced improvements in detection quality, with relative AUROC gains of up to 7.8\% over the next-best baseline.

On CoQA, the optimal aggregation is $\operatorname{INTRYGUE}_{\text{mean}}$, as this dataset contains much shorter responses (see Table~\ref{tab:response_length}). This characteristic also explains why the improvement over standard LN-Entropy is less radical: short responses frequently require logical reasoning rather than direct context copying (e.g., binary yes/no answers; see Appendix~\ref{app:coqa_fps}), so induction-head gating offers less benefit. Nevertheless, INTRYGUE still achieves the best or second-best performance in all CoQA configurations, confirming its broad applicability.

To eliminate manual aggregation selection, we also evaluate an automated variant that dynamically switches between min-max and mean using a validation-tuned response-length threshold. The results (see Appendix~\ref{app:dynamic_intrygue}) show that this length-adaptive rule matches the stronger of the two fixed aggregations, so the choice need not be made in advance.

\paragraph{Illustrative study.} Figures~\ref{fig:stability_and_samples_correction}b)-c) compare MaxEntropy and mean SinkRate predictions on CoQA (thresholds for both methods are selected on the same validation set to maximize F1-score). Quadrants show sample counts in bold and subset accuracies percentages. Notably, the bottom-left quadrant reveals that when Entropy predicts ``hallucination'' but SinkRate predicts ``grounded'', Entropy significantly underperforms (e.g., 8.1\% accuracy for Qwen). This false uncertainty is likely caused by induction head dynamics (Section~\ref{empirical_study}). INTRYGUE effectively mitigates this by lowering entropy scores wherever the SinkRate is low. While the reverse scenario (top-right quadrant)  also occurs, it affects a smaller fraction of the dataset (8–10\% for both models).

\subsection{Ablation Studies}\label{sec:ablation_studies}

\begin{figure*}[t!]
    \centering
    \includegraphics[width=\linewidth]{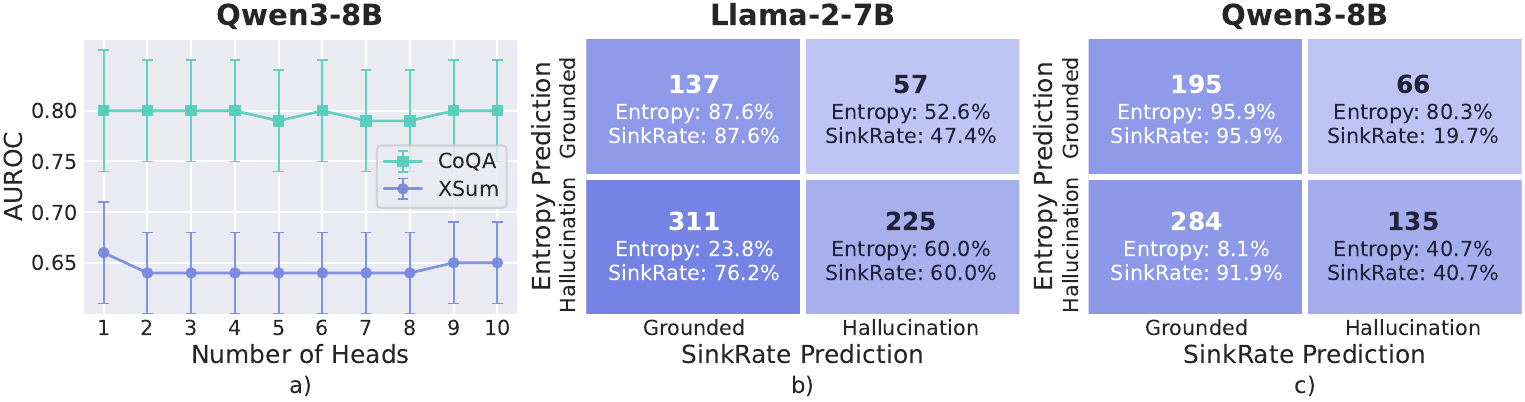}
    \caption{(a) Performance stability of $\text{INTRYGUE}_{\text{mean}}$ across different values of $k$ (number of induction heads). (b-c) MaxEntropy vs. SinkRate prediction accuracy on CoQA. The bottom-left quadrant demonstrates SinkRate's ability to correct MaxEntropy errors, a mechanism directly exploited by INTRYGUE.}
    \label{fig:stability_and_samples_correction}
\end{figure*}

\begin{figure}[h]
    \begin{minipage}{\columnwidth}
        \centering\includegraphics[width=0.6\columnwidth]{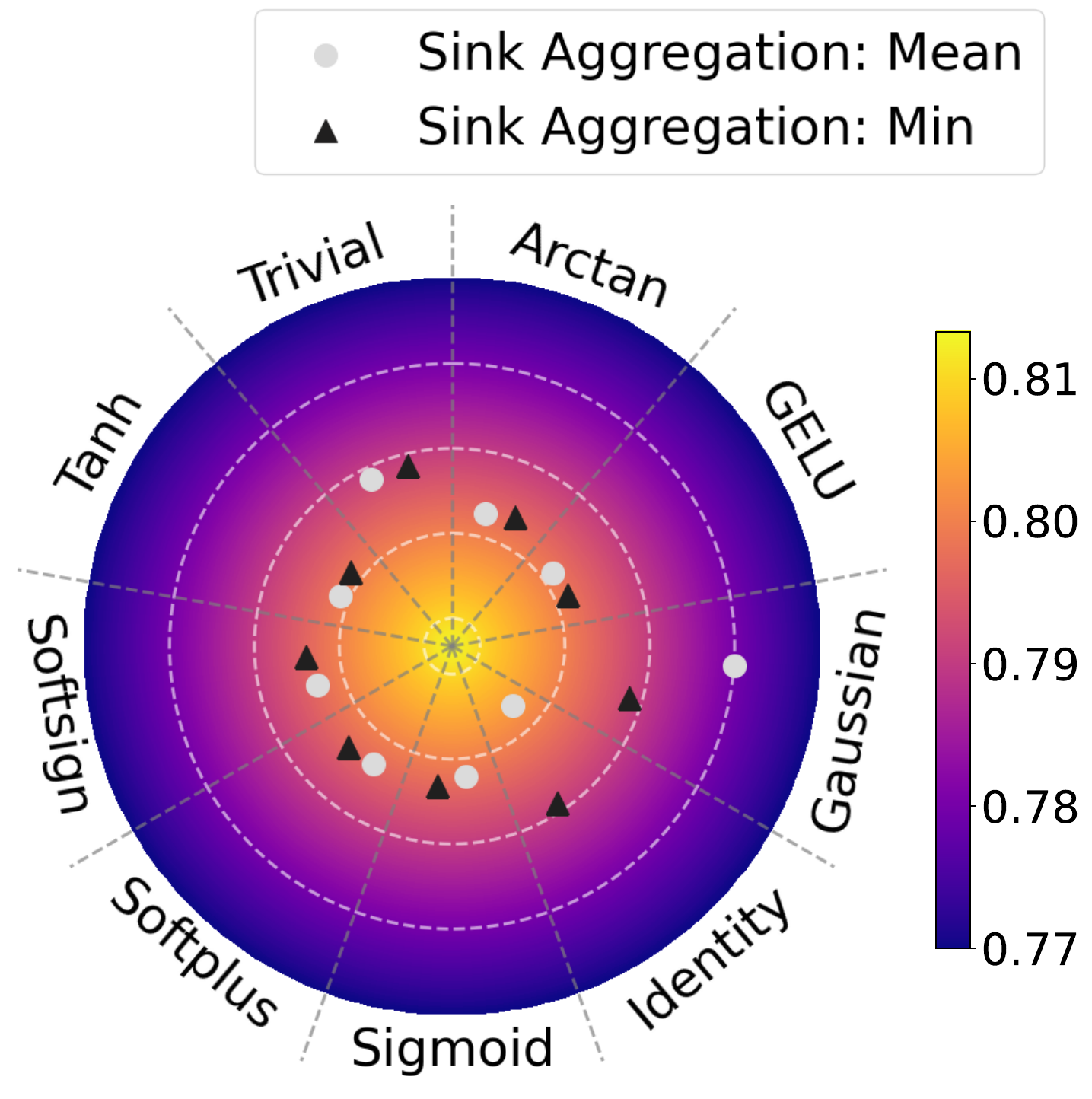}
        \caption{Impact of various non-linear transformations ($f$) on INTRYGUE performance. Higher AUROC scores are mapped closer to the center. Applying non-linear transformations does not allow for significant improvement in performance.}
        \label{fig:gating_entropy_mean}
    \end{minipage}
\end{figure}

\paragraph{Gating functions.} We tested whether applying non-linear transformations (e.g., Tanh, Softsign) to the aggregated SinkRates could improve the INTRYGUE score~\eqref{eq:intrigue}. The results in Figure~\ref{fig:gating_entropy_mean} (and Appendix~\ref{appendix:additional_results}) demonstrate that these adjustments offered no significant benefit over using the raw values directly. This confirms that our method works effectively without the need for added complexity. 

\paragraph{$\text{INTRYGUE}_{\text{min-max}}$ vs. isolated signals.} Table~\ref{tab:ablation_individual_vs_intrygue} (Appendix~\ref{appendix:additional_results}) compares our method to isolated mechanistic (SinkRate) and uncertainty (MaxEntropy) signals. As individual metric performance varies by dataset, neither signal is reliable on its own. Combining them, INTRYGUE outperforms both components in the reported settings, indicating that the two signals are complementary.

\paragraph{Number of induction heads.} Tables~\ref{tab:ragtruth} and~\ref{tab:coqa_xsum} use a validation-selected $k$ to report the peak performance of the method, which is standard practice. However, we wish to stress that INTRYGUE performs robustly even without any hyperparameter tuning. Figure~\ref{fig:stability_and_samples_correction}a) demonstrates that the average AUROC of $\text{INTRYGUE}_{\text{mean}}$ 
remains stable across $k \in \{1,\dots,10\}$. The observed run-to-run variance is inherent evaluation noise shared by all baselines (Tables~\ref{tab:ragtruth}--\ref{tab:coqa_xsum}), not a sensitivity to $k$. Consequently, INTRYGUE can be deployed with a fixed default $k$ (e.g., $k=5$) and no access to labeled validation data.

\begin{figure}[h]
    \centering
    \includegraphics[width=\linewidth]{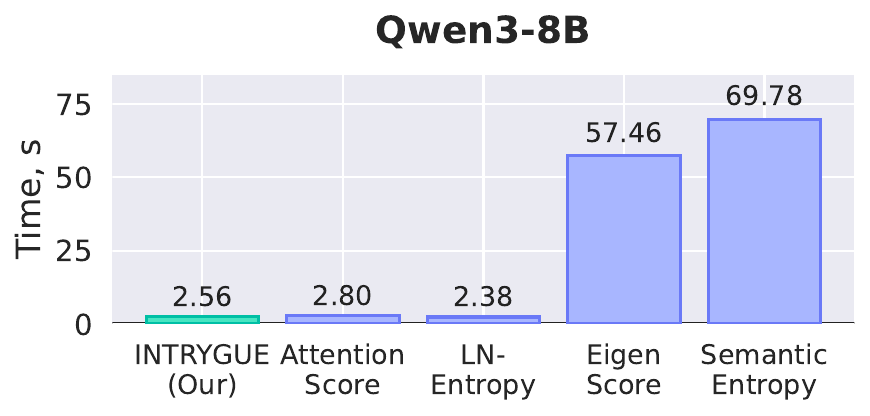}
    \caption{Computational complexity. Runtime in seconds of our method compared to information- and sampling-based baselines. Results are obtained using 20 samples from the XSum dataset with  the average over 10 runs presented.}
    \label{fig:compute_comparasion}
\end{figure}

\paragraph{Efficiency.} To highlight the efficiency of our method, we compare its computational cost with that of the LN-Entropy and Semantic Entropy baselines in Figure~\ref{fig:compute_comparasion}. INTRYGUE achieves a runtime comparable to LN-Entropy, while remaining substantially faster than the sampling-based Semantic Entropy. Similar runtimes are observed for other information-based baselines, including RAUQ ($2.74 \pm 0.03$s) and Perplexity ($2.31 \pm 0.01$s), with differences within a few tenths of a second.

\section{Related Work}\label{sec:rw}

\paragraph{Hallucination Detection in RAG Scenarios.} Approaches most relevant to our work include uncertainty-based scores as well as detectors built on context-utilization signals. Uncertainty-based methods are popular for hallucination detection in general~\cite{duan2024shifting, vazhentsev2025uncertainty, qiao2026lowest}, but they have been evaluated primarily in closed-book settings~\citep{joshi2017triviaqa, lin2022truthfulqa, kwiatkowski2019natural, wei2024measuring}, and some works argue that they do not transfer directly to RAG settings~\cite{soudani2025uncertainty}. Only a few studies adapt uncertainty estimation to RAG, and these come with limitations, such as being restricted in their application scenarios~\cite{perezuncertainty} or reliant on auxiliary models~\cite{wu2024synchronous, perezuncertainty}. Another line of work designs detectors specifically for RAG, drawing on internal signals of context utilization. Lookback Lens~\cite{chuang2024lookback}, for instance, trains a linear classifier on the ratio of attention assigned to the input context versus previously generated tokens, while LUMINA~\cite{yeh2026lumina} combines external-context utilization, measured by contrasting predictions under relevant and random contexts, with internal-knowledge utilization derived from the evolution of token predictions across layers.

\paragraph{Hallucination Detection Through Mechanistic Interpretability.} Mechanistic interpretability (MI)~\cite{ferrando2024primer} seeks to explain how individual internal components of LLMs contribute to the final prediction. MI has previously been applied to hallucination detection in several ways. \citet{hu2024lrp4rag} compare representations of external and parametric knowledge using layer-wise relevance propagation~\cite{bach2015pixel}. \citet{sunredeep} attribute hallucinations to two concurrent failures: feedforward layers overemphasize parametric knowledge in the residual stream, while copying heads~\cite{elhage2021mathematical} fail to exploit the retrieved content.

\paragraph{Summary.} Uncertainty estimation methods assign confidence scores without accounting for the extent to which the retrieved context was used, and the few adaptations to RAG incorporate this information externally, relying on auxiliary models or additional generations. Detectors based on model internals derive proxy signals of context utilization, yet they employ them to construct dedicated hallucination features rather than to explain the unreliability of established uncertainty scores. In contrast, we use internal signals to diagnose the limitations of predictive entropy rather than to bypass it: we trace its unreliability in RAG to induction-driven entropy inflation on grounded responses, and correct the score by gating it with induction-head activity.

\section{Conclusion}
In this work, we investigated a mechanistic limitation of entropy-based uncertainty estimation in RAG settings. Prior work linked induction heads to entropy-neuron activation in highly controlled settings; our results indicate that this connection extends to RAG-based decoding. We find that induction heads support correct responses and lower predictive entropy, while simultaneously activating entropy neurons that tend to counteract this reduction. We argue that this interaction helps explain why entropy-based UE can be less reliable in RAG settings.

To address this issue, we propose INTRYGUE, a training-free uncertainty estimation method that gates predictive entropy by an attention-based estimate of induction-head activity, compensating for induction-driven entropy inflation. Experiments across four benchmarks spanning question answering and summarization, evaluated on six open-source LLMs, demonstrate that our method performs on par with or better than a broad set of baselines in most evaluated settings.

Ultimately, we demonstrate that RAG uncertainty quantification could benefit from taking into account the internal mechanisms of evidence utilization. The findings motivate a promising future direction: integration of mechanistic interpretability with uncertainty estimation for building more accurate and explainable methods.

\section*{Acknowledgements}
The work was supported by the grant for research centers in the field of AI provided by the Ministry of Economic Development of the Russian Federation in accordance with the agreement 000000C313925P4F0002 and the agreement №139-10-2025-033. We are also most grateful to our colleague Ilya Kuleshov for designing Figure~\ref{fig:intro}.

\section*{Limitations}

While INTRYGUE provides a mechanistically grounded approach to uncertainty estimation in RAG, we identify the following limitations:
\begin{itemize}
    \item \textbf{Requirement of White-Box Access:} INTRYGUE relies on extracting internal attention matrices during the forward pass to compute the SinkRate and monitor induction heads. Consequently, it cannot be applied to proprietary, black-box large language models restricted behind APIs like GPT-4 or Claude.
    \item \textbf{Transformer Architecture:} Because our findings are fundamentally rooted in standard Transformer attention mechanics, the method’s applicability to alternative architectures remains an open question.
    \item \textbf{Proxy Nature of SinkRate:} SinkRate proxies induction-head activity rather than measuring hallucination directly. It is designed to capture a specific mechanistic pattern~---~context copying via induction~---~that our analysis identifies as important for faithful RAG generation (Section~3). While this makes INTRYGUE highly effective for grounding errors tied to induction dynamics, it is not intended to isolate failures arising from unrelated mechanisms (e.g., logical reasoning without context copying).
    \item \textbf{Faithfulness vs. Objective Factuality:} Our mechanistic proxy evaluates how successfully the model attends to and copies from the retrieved context. Therefore, INTRYGUE strictly measures faithfulness to the provided documents, not objective truth. If the upstream retrieval pipeline fetches false or conflicting information and the LLM successfully grounds its answer in that text, INTRYGUE will report low uncertainty, which users must not confuse with factual correctness.
\end{itemize}

\bibliography{custom}

\appendix

\begin{table*}[h!]
\centering
\caption{Hallucination statistics for CoQA and XSum tasks (generated at $t=1.0$ and annotated using GPT-4o).}
\begin{tabular}{l c c c c}
\toprule
& \multicolumn{2}{c}{CoQA} & \multicolumn{2}{c}{XSum} \\
\cmidrule(lr){2-3} \cmidrule(lr){4-5}
Model & Total & Hal. Rate (\%) & Total & Hal. Rate (\%) \\
\midrule
Mistral-7B & 1525 & 30.8 & 749 & 39.4 \\
Llama-2-7B & 741 & 34.7 & 746 & 34.9 \\
Llama-2-13B & 661 & 23.9 & 730 & 32.3 \\
Llama-3.1-8B & 691 & 22.1 & 650 & 28.9 \\
Qwen3-8B & 698 & 14.6 & 750 & 24.3 \\
gemma-3-4b & 703 & 22.1 & 650 & 31.3 \\
\bottomrule
\end{tabular}
\label{tab:dataset_stats}
\end{table*}

\begin{table*}[t]
\centering
\caption{Median response length (words) and central 95\% interval $[P_{2.5}, P_{97.5}]$ for each model--dataset pair.}
\label{tab:response_length}
\resizebox{\textwidth}{!}{%
\begin{tabular}{lcccccccc}
\toprule
Model & \multicolumn{2}{c}{CoQA} & \multicolumn{2}{c}{XSum} & \multicolumn{2}{c}{MS MARCO} & \multicolumn{2}{c}{CNN / DM} \\
\cmidrule(lr){2-3} \cmidrule(lr){4-5} \cmidrule(lr){6-7} \cmidrule(lr){8-9}
 & Median & $[P_{2.5}, P_{97.5}]$ & Median & $[P_{2.5}, P_{97.5}]$ & Median & $[P_{2.5}, P_{97.5}]$ & Median & $[P_{2.5}, P_{97.5}]$ \\
\midrule
Mistral-7B & 4 & [1, 17] & 43 & [18, 138.9] & 93 & [7, 204] & 90 & [41.4, 163.6] \\
Llama-2-7B & 5 & [1, 17] & 49 & [30, 113.8] & 168 & [68, 319.3] & 78 & [34, 120] \\
Llama-2-13B & 5 & [1, 15] & 55 & [37, 77] & 139 & [43, 258.3] & 77 & [36.4, 112.6] \\
Llama-3.1-8B & 3 & [1, 14] & 36 & [24.2, 54] & --- & --- & --- & --- \\
Qwen3-8B & 2 & [1, 13] & 32.5 & [21, 45.3] & --- & --- & --- & --- \\
gemma-3-4b & 3 & [1, 14] & 36 & [24, 49] & --- & --- & --- & --- \\
\bottomrule
\end{tabular}%
}
\end{table*}

\section{Datasets}\label{sec:appendix_datasets}

To evaluate the effectiveness of our approach across diverse RAG scenarios, we conduct experiments on four established benchmarks spanning both question answering (QA) and summarization. These datasets present varying levels of difficulty for context grounding and hallucination detection:

\begin{itemize}
    \item \textbf{RAGTruth (MS MARCO \& CNN/DM):} We utilize the long-form QA (MS MARCO) and standard summarization (CNN/DM) tasks from the RAGTruth benchmark~\cite{wu2023ragtruth}. These subsets provide high-quality annotated data specifically designed for evaluating hallucinations in retrieval-augmented generation.
    
    \item \textbf{CoQA:} To test uncertainty quantification in interactive, context-dependent settings, we evaluate on the Conversational Question Answering (CoQA) dataset~\cite{reddy2019coqa}, which requires models to ground responses over multi-turn dialogues. To construct our evaluation instances for this task, we generated model responses at a temperature of $t=1.0$ and subsequently annotated them for hallucinations using GPT-4o.
    
    \item \textbf{XSum:} Finally, we include the Extreme Summarization dataset (XSum)~\cite{narayan2018don}. Because XSum requires highly abstractive, single-sentence summaries, models are notoriously prone to hallucination here, providing a rigorous stress-test for our uncertainty estimator. As with CoQA, the summaries used for evaluation were generated at $t=1.0$ and automatically annotated using GPT-4o. 
\end{itemize}

The statistics of the CoQA and XSum datasets are provided in Table~\ref{tab:dataset_stats}. 

To ensure the quality of LLM-based annotation, we conducted the following study. Three team members with at least upper-intermediate English proficiency independently annotated 100 randomly selected samples from CoQA. We first measured inter-annotator consistency between human annotators. The resulting human–human agreement statistics are shown in Table~\ref{tab:annotator_agreement}. We then selected the samples for which all annotators reached consensus and treated these labels as the ground-truth hallucination annotations. To evaluate GPT as an automatic annotator, we prompted the model using five instruction variants, including both zero-shot and few-shot formulations, and compared its annotations against the human labels (see Table~\ref{tab:human_vs_gpt} for consistency metrics). We can conclude that average human–AI agreement accuracy (0.808) closely matches the average human–human agreement accuracy (0.807); these results indicate that the automatic annotations used in our work provide a practically reliable approximation for this task.

To further address potential concerns regarding annotation quality, we repeated the main experiments for Gemma-3-4B on the AI-annotated CoQA dataset using hallucination labels generated by Gemini 3.1 Pro Preview. Importantly, we used exactly the same set of Gemma-3-4B responses and preserved all other experimental procedures unchanged. Among the 703 newly annotated CoQA samples, 641 labels matched the original annotations (91\% agreement rate). We additionally investigated the impact of the remaining 62 mismatched samples on evaluation performance by re-running our method together with the strongest-performing baselines under both annotation sets. As shown in Table~\ref{tab:before_after_methods}, the resulting performance metrics remain closely aligned with those obtained using the original annotations.

\begin{table}[t]
\centering
\caption{Inter-annotator agreement on CoQA.}
\label{tab:annotator_agreement}
\small
\begin{tabular}{lcc}
\toprule
Annotator Pair & Accuracy & Pearson-$r$ \\
\midrule
A vs B & 0.783 & 0.600 \\
A vs C & 0.812 & 0.631 \\
B vs C & 0.826 & 0.653 \\
\midrule
Mean & 0.807 & 0.628 \\
Std & 0.022 & 0.027 \\
\bottomrule
\end{tabular}
\end{table}

\begin{table*}[!htp]
\centering
\caption{Classification metrics of GPT-4o annotation for CoQA with human labels considered as true. The table shows metric scores for different prompt variants and the average score across all variants.}
 \scriptsize
\resizebox{\textwidth}{!}{
    \begin{tabular}{llcccccc}
      \hline
      \multicolumn{2}{c}{Prompt number} & 1 & 2 & 3 & 4 & 5 & Average \\
      \hline
      \multirow{3}{*}{CoQA} & Accuracy ($\uparrow$) & 0.809 ± 0.017 & 0.861 ± 0.015 & 0.742 ± 0.003 & 0.795 ± 0.009 & 0.831 ± 0.025 & 0.808\\
      & Precision ($\uparrow$) & 0.849 ± 0.021 & 0.911 ± 0.007 & 0.771 ± 0.003 & 0.828 ± 0.011 & 0.860 ± 0.012 & 0.844\\
      & Recall ($\uparrow$) & 0.871 ± 0.004 & 0.877 ± 0.019 & 0.877 ± 0.013 & 0.877 ± 0.005 & 0.893 ± 0.027 & 0.879\\
      \hline
    \end{tabular}
    \newline
    \vspace*{0.4cm}
    \newline
}
\label{tab:human_vs_gpt}
\end{table*}

\begin{table}[t]
\centering
\caption{Performance metrics using GPT-4o annotation (before) and Gemini-3.1 Pro annotation (after) as the ground truth hallucination labels.}
\label{tab:before_after_methods}
\begin{tabular}{lcc}
\toprule
Method & Before & After \\
\midrule
$\text{INTRYGUE}_{\text{min-max}}$ & $0.73$ \scriptsize{$\pm$ 0.05} & $0.71$ \scriptsize{$\pm 0.03$} \\
$\text{INTRYGUE}_{\text{mean}}$ & $0.76$ \scriptsize{$\pm 0.04$} & $0.76$ \scriptsize{$\pm 0.05$} \\
LN-Entropy & $0.75$  \scriptsize{$\pm 0.04$} & $0.73$ \scriptsize{$\pm 0.05$} \\
EigenScore & $0.71$ \scriptsize{$\pm 0.05$} & $0.67$ \scriptsize{$\pm 0.03$} \\
Semantic Entropy & $0.70$  \scriptsize{$\pm 0.05$} & $0.69$ \scriptsize{$\pm 0.01$} \\
Lookback Lens & $0.62$ \scriptsize{$\pm 0.05$} & $0.61$ \scriptsize{$\pm 0.12$} \\
\bottomrule
\end{tabular}
\end{table}

\section{Additional results} \label{appendix:additional_results}
\subsection{Mechanistic Experiments Ablations}
\begin{figure}[t!]
    \centering
    \includegraphics[width=\linewidth]{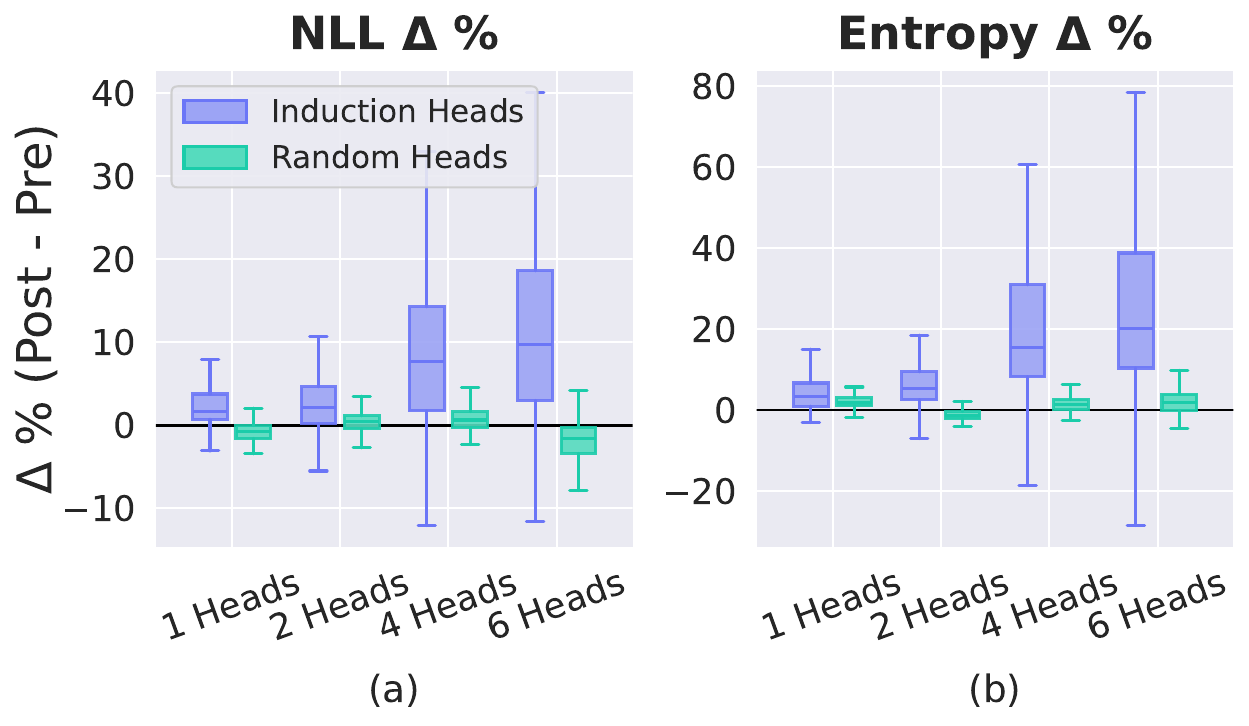} 
    \caption{Ablation over the number of induction heads for the experiment in Figure~\ref{fig:induction_heads_causal_nll}.}
    \label{fig:num_heads_ablation} 
\end{figure}

To assess the robustness of the effects shown in Figures~\ref{fig:induction_heads_causal_nll} and~\ref{fig:induction_heads_vs_neurons}, we run additional ablations on Llama-3.1-8B-Instruct.
\paragraph{Induction Heads Pull Confidence Up: ablation on the number of induction heads.} In section~\ref{sec:pull_confidence_up}, we report results for the top-5 induction heads. Here, we extend the analysis to subsets of various size (top-\{1, 2, 4, 6\}) to test whether the trend is sensitive to this choice. The results are shown in Figure~\ref{fig:num_heads_ablation}. Across all settings, ablating induction heads produces consistently larger shifts in NLL and entropy, which further supports our claims.

\begin{figure}[t!]
    \centering
    \includegraphics[width=\linewidth]{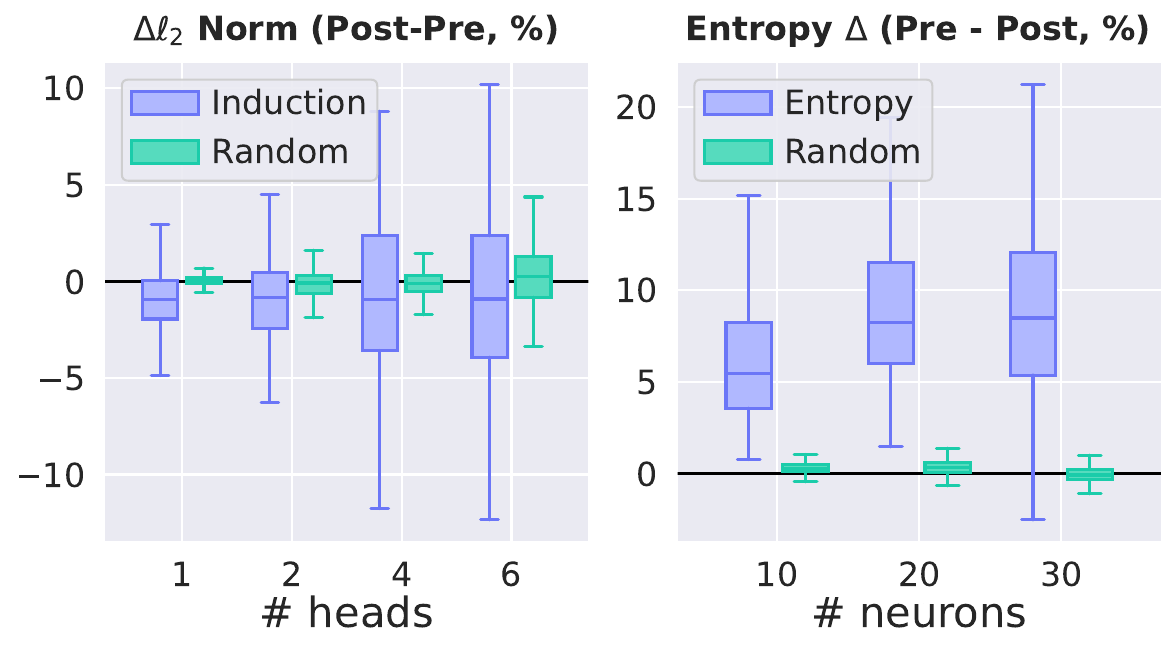} 
    \caption{Ablation over the number of induction heads and the number of entropy neurons for the experiment in Figure~\ref{fig:induction_heads_vs_neurons}.}
    \label{fig:num_neurons_ablation} 
\end{figure}

\paragraph{..and Push It Back Down: ablations on the number of induction heads and entropy neurons.} Additional ablations suggest that the findings in Section~\ref{sec:and_push_back_down} are also robust to the exact number of components considered. As shown in Figure~\ref{fig:num_neurons_ablation}, similar trends are observed when varying the number of induction heads and when ablating the top 10, 20, or 30 entropy neurons.

\subsection{Analysis of Performance on CoQA}\label{app:coqa_fps}

Compared to the other datasets, CoQA contains substantially shorter responses (Table~\ref{tab:response_length}). These responses more often rely on concise reasoning (e.g., binary yes/no answers; see Table~\ref{tab:binary_response_frac}) rather than explicit context copying. In such cases, induction-related signals can be weaker, so SinkRate may stay relatively high even for correct answers, which can increase uncertainty estimates.

To examine this effect, we measured false positive rates on binary yes/no responses in CoQA for LN-Entropy and INTRYGUE. The rates are 0.38 vs.\ 0.40 for Mistral-7B, and 0.31 vs.\ 0.43 for Llama-3.1-8B, respectively. These findings are consistent with our hypothesis that some correct responses not driven by direct context copying receive elevated uncertainty under INTRYGUE. This likely helps explain why INTRYGUE’s gains on CoQA are more modest than on the other considered datasets.

\begin{table}[t]
\centering
{\small \setlength{\tabcolsep}{4pt}
\renewcommand{\arraystretch}{0.8}
\caption{Fraction of binary (yes/no) responses per model--dataset pair. A response is counted as binary if, after lowercasing and stripping punctuation, it is a yes/no variant (e.g., \textit{yes}, \textit{no.}, \textit{Nope}, \textit{yeah}).}
\label{tab:binary_response_frac}
\begin{tabular}{lcccc}
\toprule
Model & CoQA & XSum & MS MARCO & CNN/DM \\
\midrule
Mistral-7B & 8.2\% & 0\% & 0\% & 0\% \\
Llama-2-7B & 12.7\% & 0\% & 0\% & 0\% \\
Llama-2-13B & 10.0\%  & 0\% & 0\% & 0\% \\
Llama-3.1-8B & 9.1\% & 0\% & --- & --- \\
Qwen3-8B & 11.3\% & 0\% & --- & --- \\
gemma-3-4b & 9.8\% & 0\% & --- & --- \\
\bottomrule
\end{tabular}}
\end{table}

\subsection{Dynamic INTRYGUE}\label{app:dynamic_intrygue}

Because the preferred aggregation depends on response length, requiring practitioners to choose between $\operatorname{INTRYGUE}_{\text{mean}}$ and $\operatorname{INTRYGUE}_{\text{min-max}}$ a priori limits practical deployment.
To remove this choice, we introduce $\operatorname{INTRYGUE}_{\text{dynamic}}$, which selects mean or min-max aggregation by thresholding response length, with the threshold tuned on a validation set.
As shown in Table~\ref{tab:intrygue_dynamic}, $\operatorname{INTRYGUE}_{\text{dynamic}}$ matches or closely approaches the stronger of the two fixed variants across models and datasets, and in several settings even exceeds both.
These results indicate that a single length-adaptive rule recovers the benefits of aggregation specialization without committing to one aggregation in advance.

\begin{table*}[h!]
\centering
\caption{AUROC scores for $\text{INTRYGUE}_{\text{min-max}}$, $\text{INTRYGUE}_{\text{mean}}$, and $\text{INTRYGUE}_{\text{dynamic}}$ on CoQA, MS MARCO, CNN/DM, and XSum. Bold and underline indicate the best and second-best results among the three variants, respectively. All results are averaged over 5 runs.}
{\small \setlength{\tabcolsep}{4pt}
\renewcommand{\arraystretch}{0.8}
\begin{tabular}{l|cccccc}\toprule
                   & \multicolumn{6}{c}{CoQA} \\
                   \midrule
                   & Mistral-7B & Llama-2-7B & Llama-2-13B & Llama-3.1-8B & Qwen3-8B & gemma-3-4b \\
        \midrule
$\text{INTRYGUE}_{\text{min-max}}$ & 0.79 \scriptsize{$\pm$ 0.04} & 0.76 \scriptsize{$\pm$ 0.04} & 0.72 \scriptsize{$\pm$ 0.05} & 0.78 \scriptsize{$\pm$ 0.08} & 0.77 \scriptsize{$\pm$ 0.07} & 0.73 \scriptsize{$\pm$ 0.05} \\
$\text{INTRYGUE}_{\text{mean}}$    & \underline{0.82 \scriptsize{$\pm$ 0.04}} & \textbf{0.79 \scriptsize{$\pm$ 0.03}} & \textbf{0.81 \scriptsize{$\pm$ 0.03}} & \underline{0.81 \scriptsize{$\pm$ 0.07}} & \textbf{0.80 \scriptsize{$\pm$ 0.06}} & \textbf{0.76 \scriptsize{$\pm$ 0.04}} \\
$\text{INTRYGUE}_{\text{dynamic}}$ & \textbf{0.83 \scriptsize{$\pm$ 0.04}} & \textbf{0.79 \scriptsize{$\pm$ 0.02}} & \underline{0.80 \scriptsize{$\pm$ 0.03}} & \textbf{0.82 \scriptsize{$\pm$ 0.07}} & \textbf{0.80 \scriptsize{$\pm$ 0.06}} & \textbf{0.76 \scriptsize{$\pm$ 0.05}} \\
\midrule
                   & \multicolumn{3}{c}{MS MARCO} & \multicolumn{3}{c}{CNN/DM} \\
                   \midrule
                   & Mistral-7B & Llama-2-7B & Llama-2-13B & Mistral-7B & Llama-2-7B & Llama-2-13B \\
        \midrule
$\text{INTRYGUE}_{\text{min-max}}$ & \underline{0.77 \scriptsize{$\pm$ 0.03}} & \underline{0.72 \scriptsize{$\pm$ 0.04}} & \textbf{0.73 \scriptsize{$\pm$ 0.04}} & \underline{0.69 \scriptsize{$\pm$ 0.03}} & 0.60 \scriptsize{$\pm$ 0.03} & \textbf{0.60 \scriptsize{$\pm$ 0.08}} \\
$\text{INTRYGUE}_{\text{mean}}$    & 0.71 \scriptsize{$\pm$ 0.06} & 0.67 \scriptsize{$\pm$ 0.02} & 0.67 \scriptsize{$\pm$ 0.04} & 0.65 \scriptsize{$\pm$ 0.03} & \textbf{0.62 \scriptsize{$\pm$ 0.08}} & \underline{0.58 \scriptsize{$\pm$ 0.03}} \\
$\text{INTRYGUE}_{\text{dynamic}}$ & \textbf{0.84 \scriptsize{$\pm$ 0.03}} & \textbf{0.74 \scriptsize{$\pm$ 0.03}} & \textbf{0.73 \scriptsize{$\pm$ 0.05}} & \textbf{0.70 \scriptsize{$\pm$ 0.04}} & \textbf{0.62 \scriptsize{$\pm$ 0.03}} & 0.57 \scriptsize{$\pm$ 0.07} \\
\midrule
                   & \multicolumn{6}{c}{XSum} \\
                   \midrule
                   & Mistral-7B & Llama-2-7B & Llama-2-13B & Llama-3.1-8B & Qwen3-8B & gemma-3-4b \\
        \midrule
$\text{INTRYGUE}_{\text{min-max}}$ & \textbf{0.73 \scriptsize{$\pm$ 0.07}} & \textbf{0.66 \scriptsize{$\pm$ 0.07}} & \textbf{0.73 \scriptsize{$\pm$ 0.04}} & \textbf{0.66 \scriptsize{$\pm$ 0.02}} & \textbf{0.68 \scriptsize{$\pm$ 0.06}} & \textbf{0.62 \scriptsize{$\pm$ 0.06}} \\
$\text{INTRYGUE}_{\text{mean}}$    & \underline{0.70 \scriptsize{$\pm$ 0.03}} & 0.63 \scriptsize{$\pm$ 0.03} & 0.62 \scriptsize{$\pm$ 0.03} & 0.64 \scriptsize{$\pm$ 0.03} & 0.66 \scriptsize{$\pm$ 0.04} & 0.58 \scriptsize{$\pm$ 0.06} \\
$\text{INTRYGUE}_{\text{dynamic}}$ & \textbf{0.73 \scriptsize{$\pm$ 0.07}} & \underline{0.64 \scriptsize{$\pm$ 0.06}} & \underline{0.63  \scriptsize{$\pm$ 0.06}}  & \underline{0.65 \scriptsize{$\pm$ 0.02}} & \underline{0.67 \scriptsize{$\pm$ 0.05}} & \underline{0.61 \scriptsize{$\pm$ 0.06}} \\
\bottomrule
\end{tabular}}
\label{tab:intrygue_dynamic}
\end{table*}

\subsection{Gating Functions}
\label{appendix:gating}
\begin{figure*}
\centering
\begin{subfigure}{0.49\textwidth}
    \centering
    \includegraphics[width=0.7\columnwidth]{figures/coqa_Qwen3-8B_calibrated_mean_entropy_abalation.pdf}
    \caption{Qwen3-8B | CoQA}
\end{subfigure}
\hfill
\begin{subfigure}{0.49\textwidth}
    \centering
    \includegraphics[width=0.7\columnwidth]{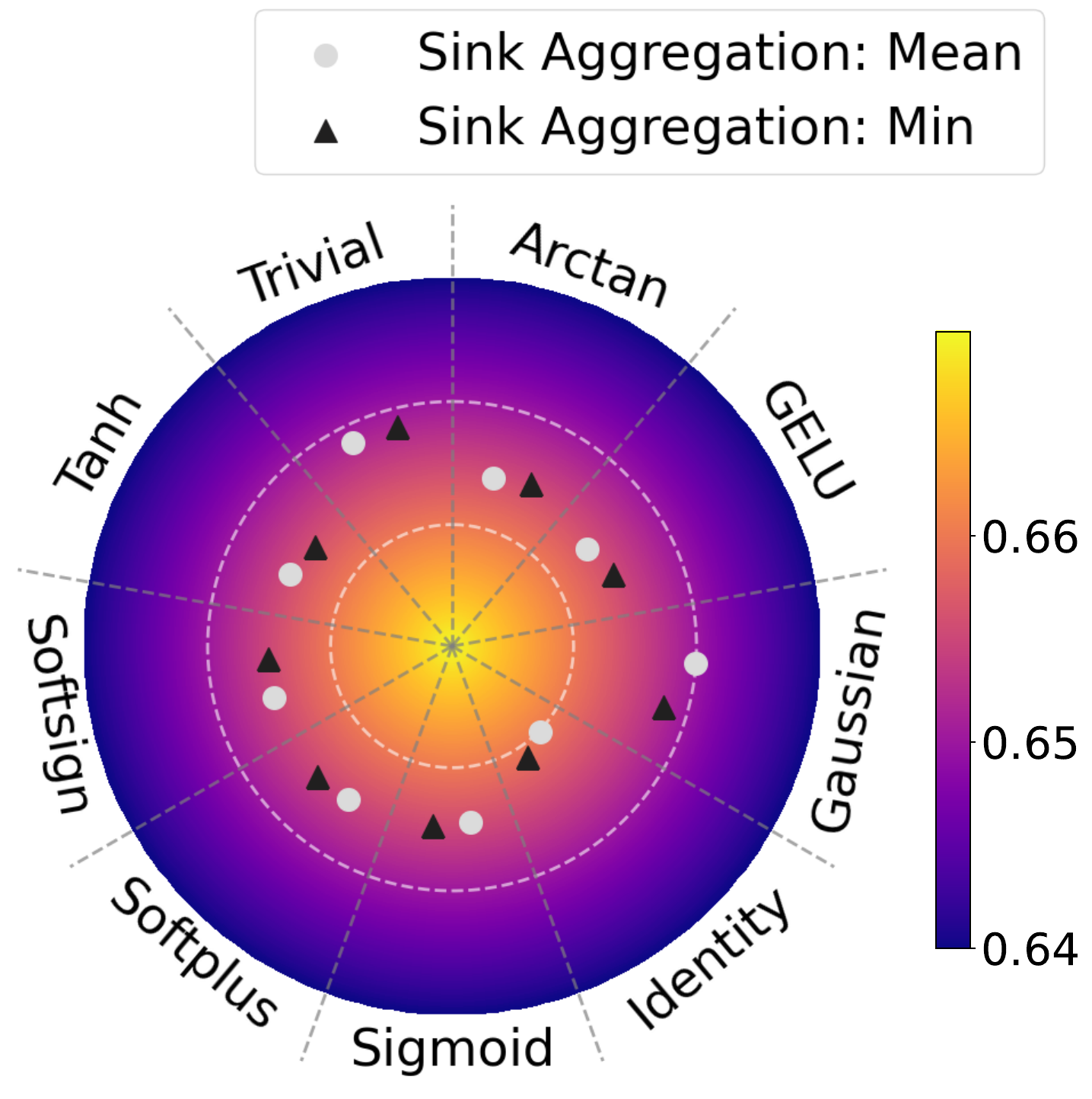}
    \caption{Qwen3-8B | XSum}
\end{subfigure}
\begin{subfigure}{0.49\textwidth}
    \centering
    \includegraphics[width=0.7\columnwidth]{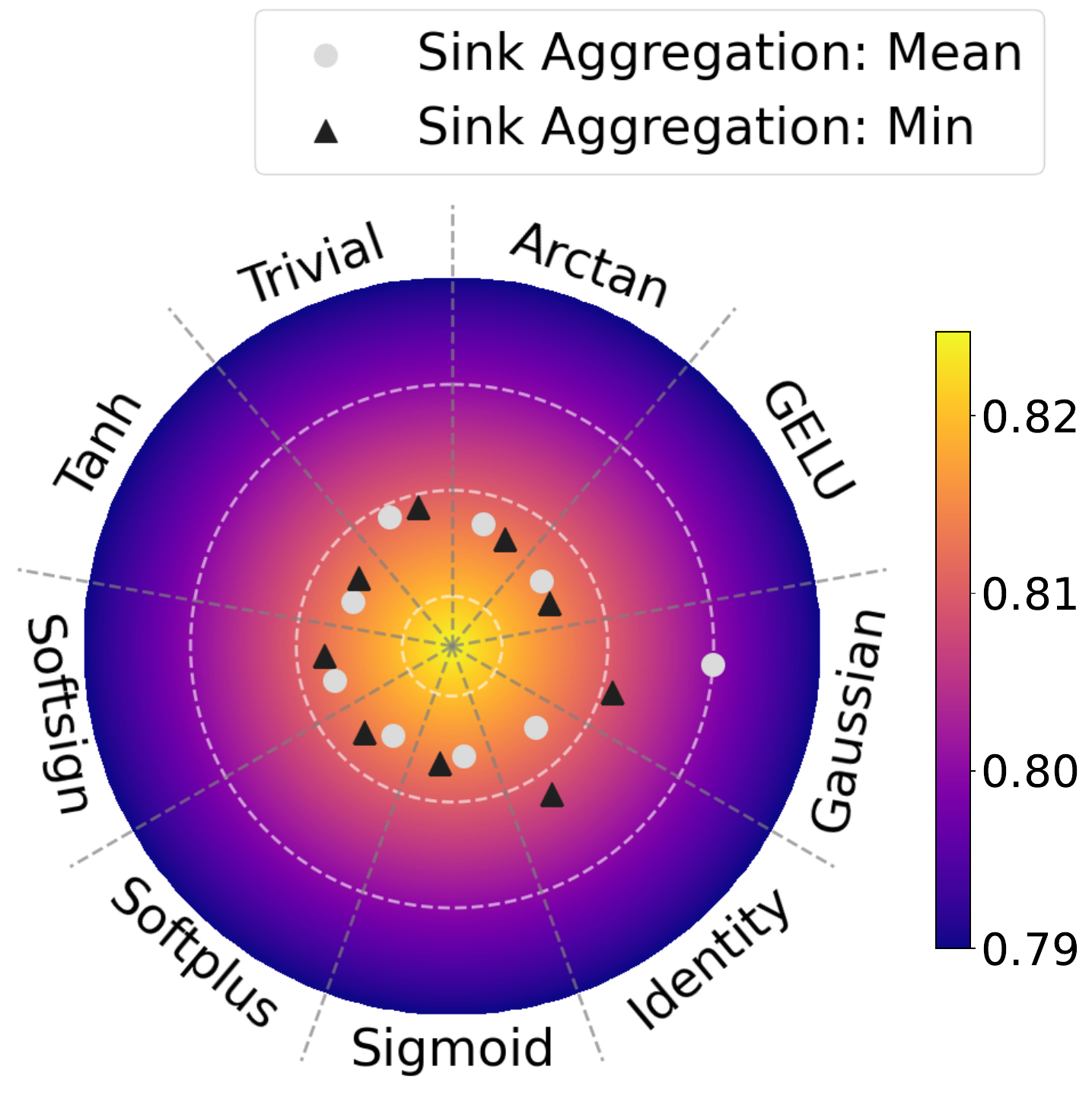}
    \caption{Llama-3.1-8B-Instruct | CoQA}
\end{subfigure}
\hfill
\begin{subfigure}{0.49\textwidth}
    \centering
    \includegraphics[width=0.7\columnwidth]{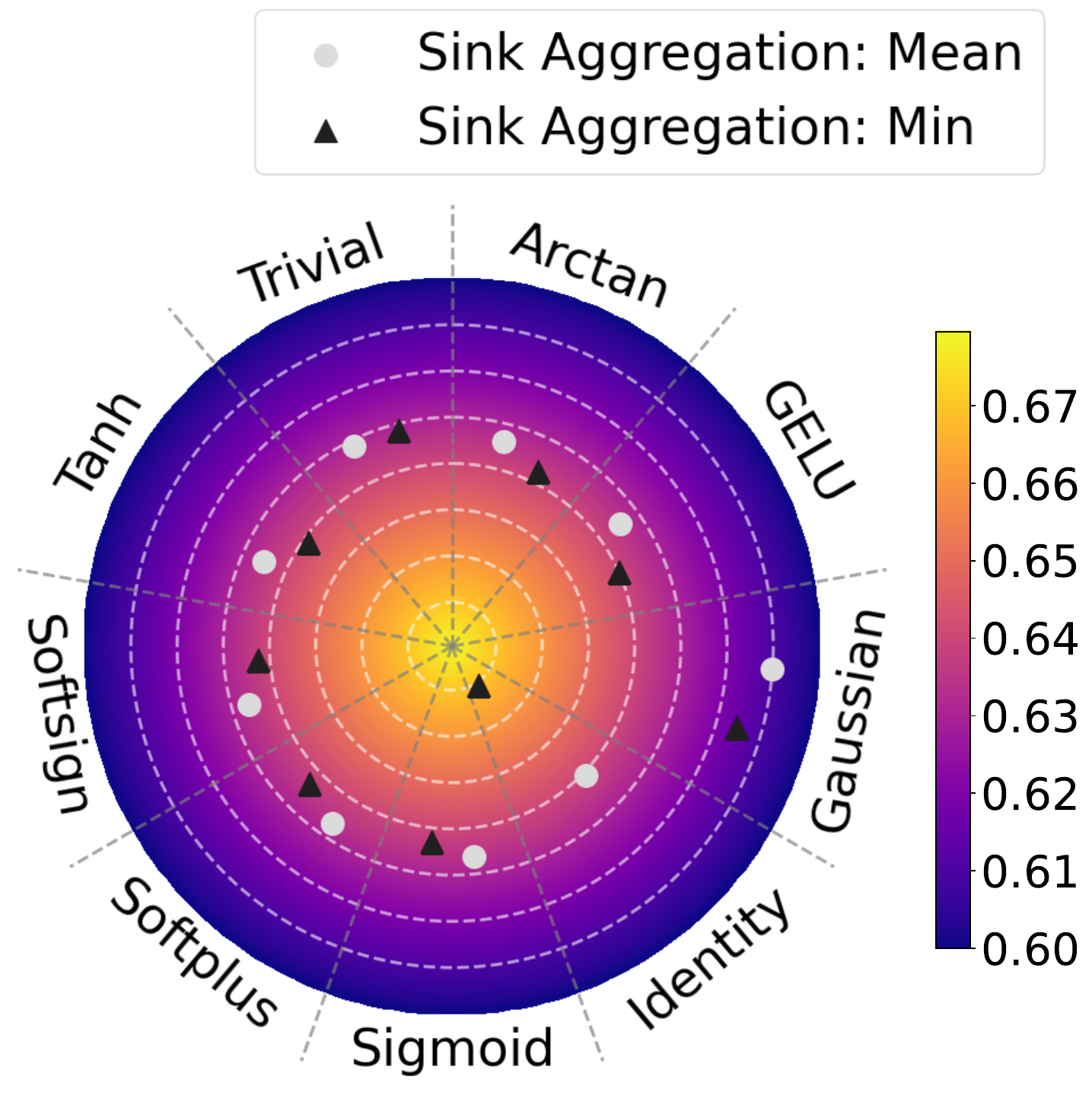}
    \caption{Llama-3.1-8B-Instruct | XSum}
\end{subfigure}
\caption{\textbf{Mean entropy aggregation.} Ablation of non-linear mappings that project the SinkRate to the $[0,1]$ interval before multiplying it with entropy (Eq.~\ref{eq:intrigue}). Trivial denotes that the SinkRate is set to 1, while Identity indicates that no mapping is applied. The heatmap shows AUC scores, where the center corresponds to AUC=1. Mappings are applied after Sink aggregation, and results are reported for both Mean and Min aggregations.}
\label{fig:additional_gating_entropy_mean}
\end{figure*}

\begin{figure*}
\centering
\begin{subfigure}{0.49\textwidth}
    \centering
    \includegraphics[width=0.7\columnwidth]{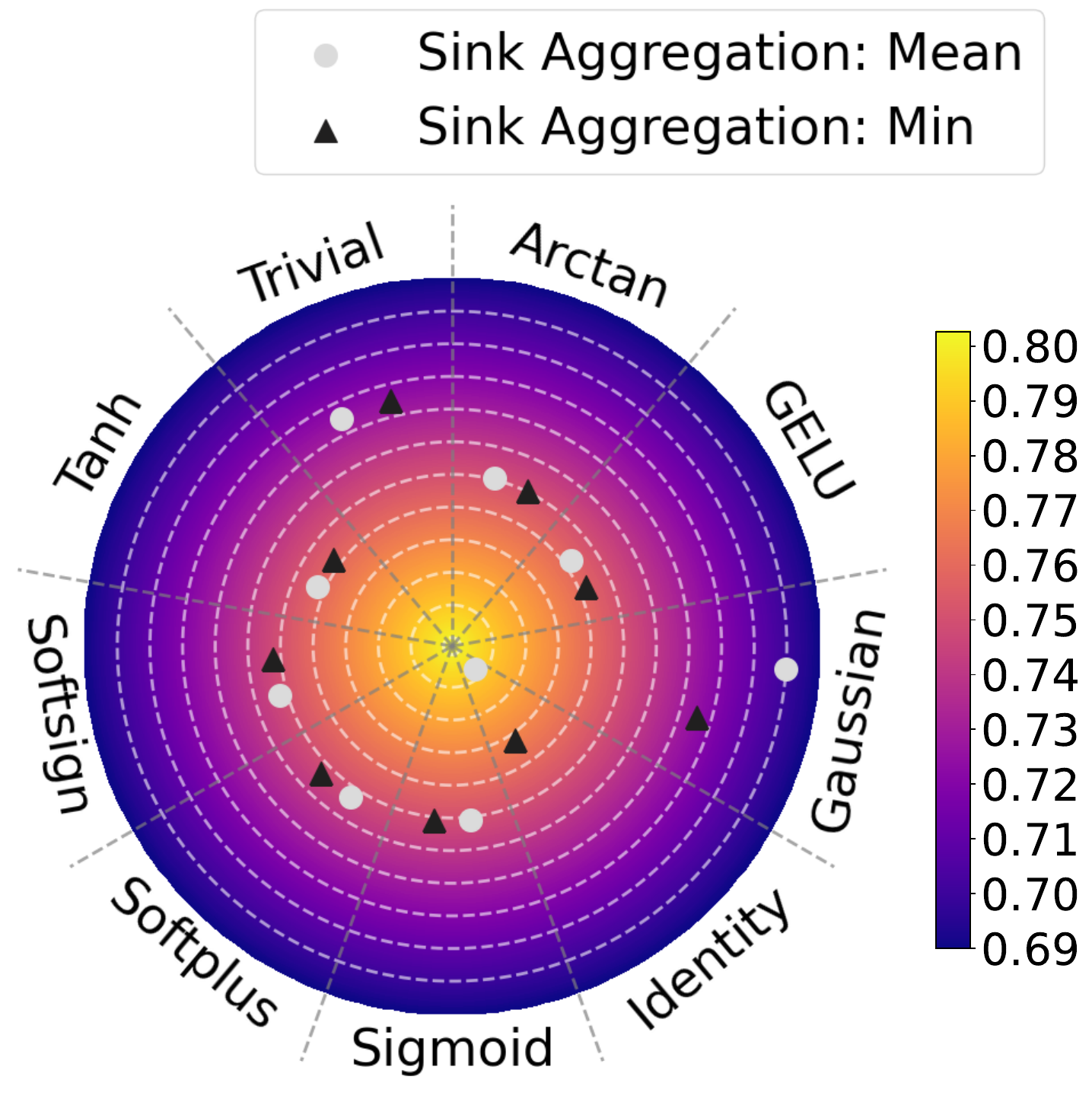}
    \caption{Qwen3-8B | CoQA}
\end{subfigure}
\hfill
\begin{subfigure}{0.49\textwidth}
    \centering
    \includegraphics[width=0.7\columnwidth]{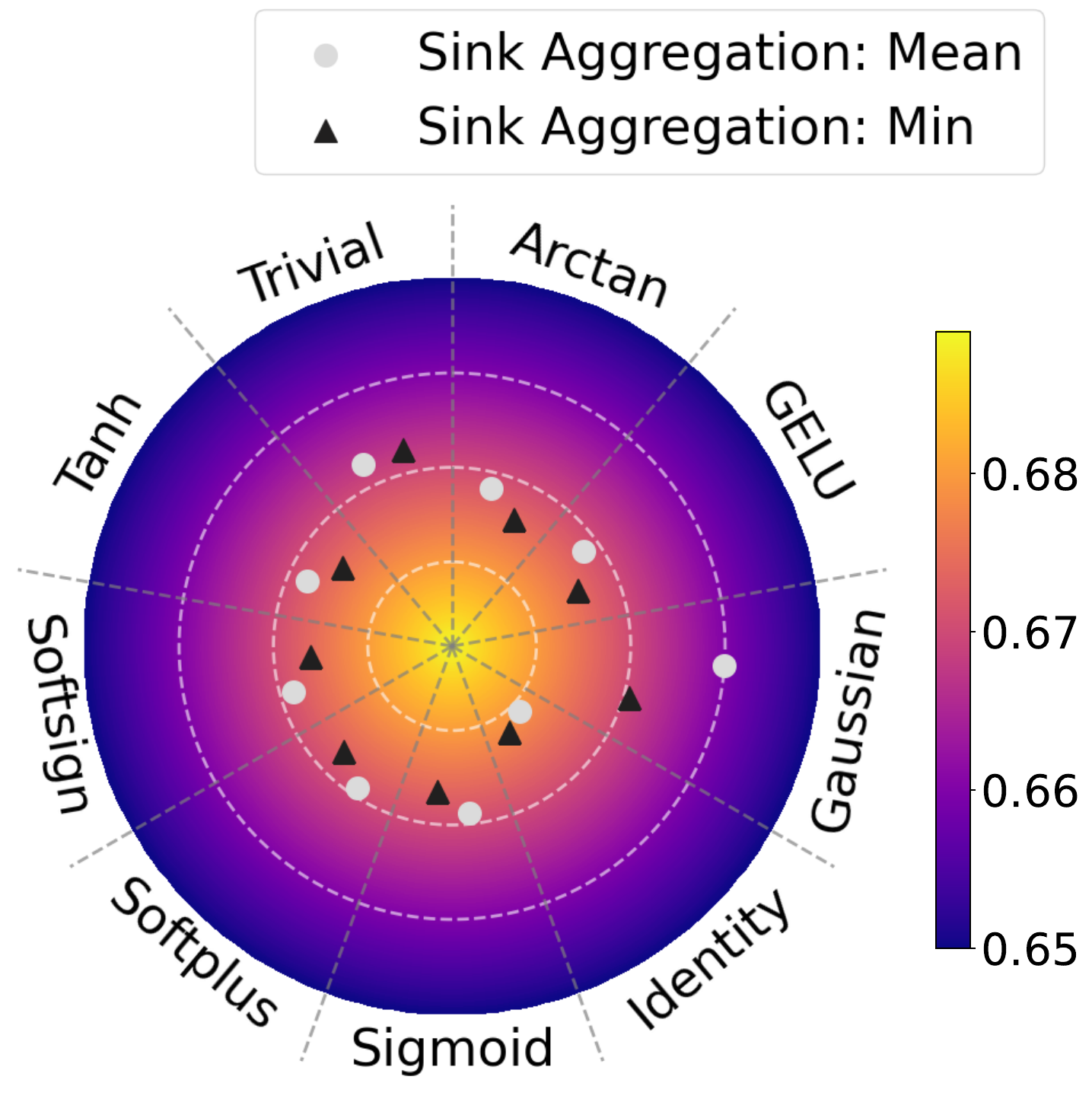}
    \caption{Qwen3-8B | XSum}
\end{subfigure}
\hfill
\begin{subfigure}{0.49\textwidth}
    \centering
    \includegraphics[width=0.7\columnwidth]{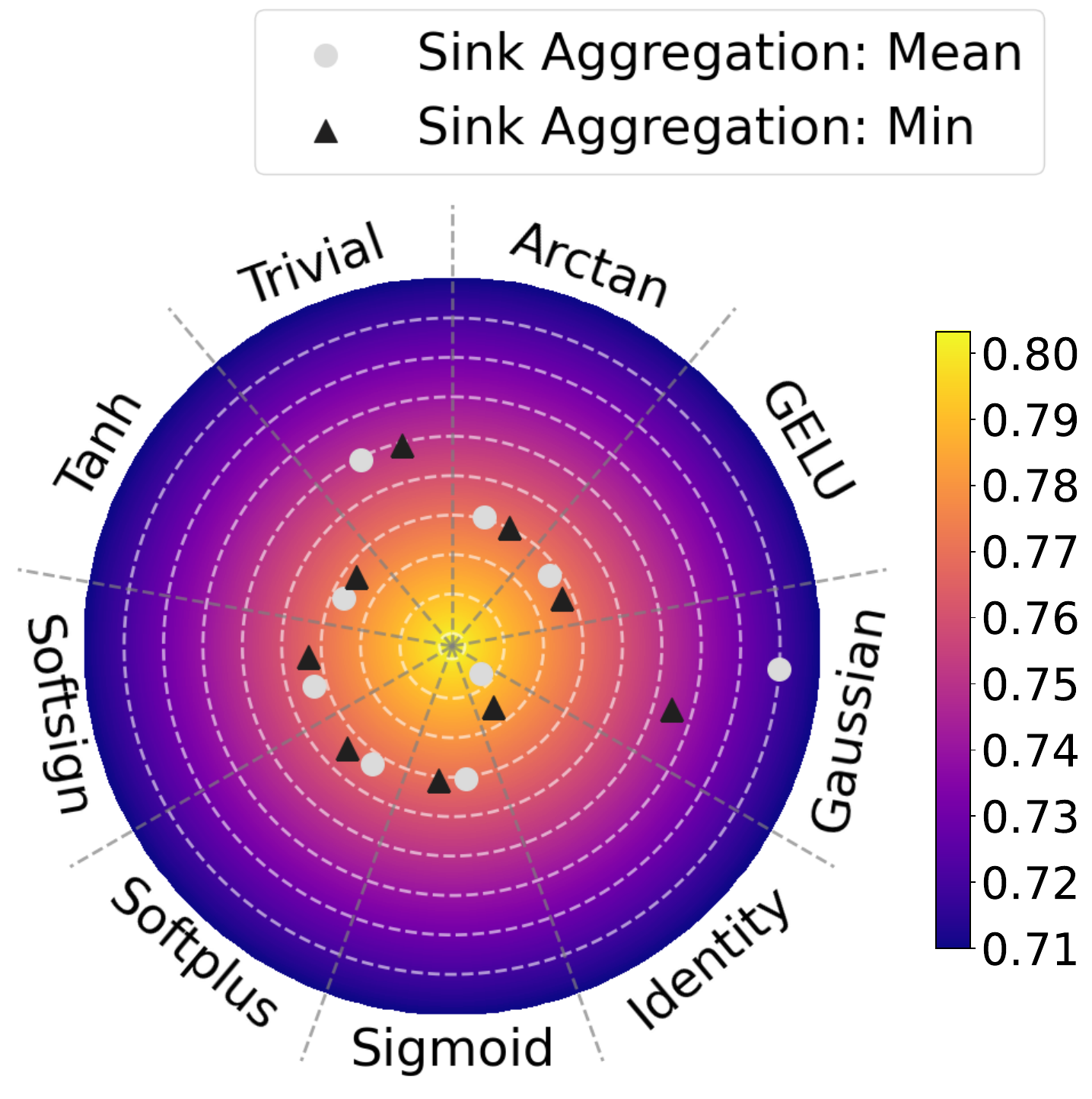}
    \caption{Llama-3.1-8B-Instruct | CoQA}
\end{subfigure}
\hfill
\begin{subfigure}{0.49\textwidth}
    \centering
    \includegraphics[width=0.7\columnwidth]{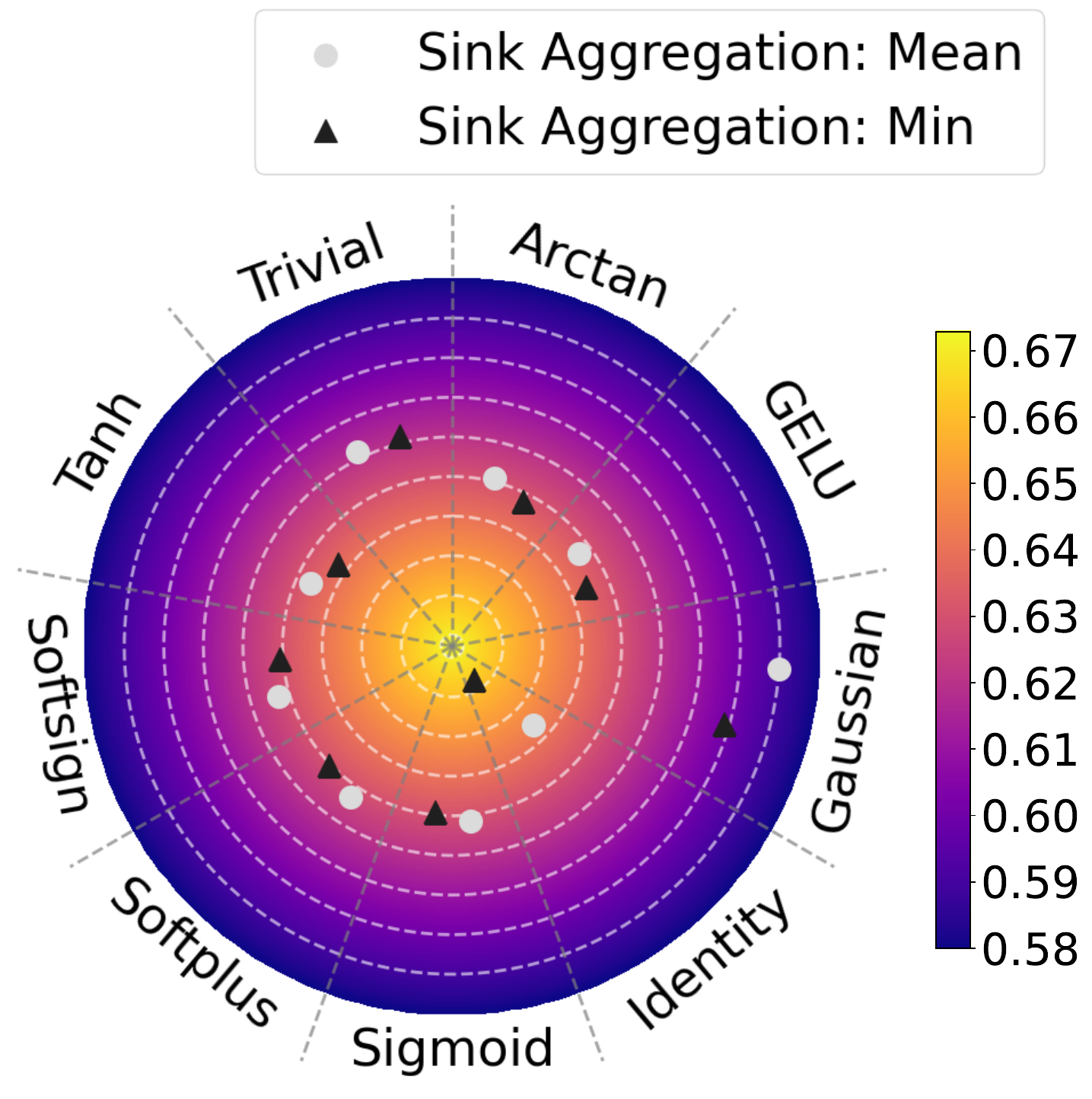}
    \caption{Llama-3.1-8B-Instruct | XSum}
\end{subfigure}

\caption{\textbf{Max entropy aggregation.} Ablation of non-linear mappings that project the SinkRate to the $[0,1]$ interval before multiplying it with entropy (Eq.~\ref{eq:intrigue}). Trivial denotes that the SinkRate is set to 1, while Identity indicates that no mapping is applied. The heatmap shows AUC scores, where the center corresponds to AUC=1. Mappings are applied after Sink aggregation, and results are reported for both Mean and Min aggregations.}
\label{fig:additional_gating_entropy_max}
\end{figure*}

The additional results on applying non-linear mappings to the SinkRate are demonstrated in  Figures~\ref{fig:additional_gating_entropy_mean} and ~\ref{fig:additional_gating_entropy_max}. They confirm that additional transformations are uncecessary for our method.

\begin{table*}[htbp]
\centering
{\small
    \caption{Performance comparison of minimum Sink Rate, maximum Entropy, and INTRYGUE across selected models on the CoQA and MS MARCO datasets. INTRYGUE outperforms both individual scores, demonstrating that combining uncertainty estimates with mechanistic signals yields superior detection compared to relying on either signal alone.} \label{tab:ablation_individual_vs_intrygue}
    \begin{tabular}{lccc}
        \toprule
        \multicolumn{4}{c}{CoQA} \\
        \midrule
        Method & Llama-2-13B & Qwen3-8B & Gemma-3-4B \\
        \midrule
        SinkRate (min)             & $0.64 \pm 0.03$ & $0.57 \pm 0.04$ & $0.63 \pm 0.04$ \\
        MaxEntropy       &  $0.65 \pm 0.06$ &	$0.73 \pm 0.05$ &	$0.70 \pm 0.07$ \\
        $\text{INTRYGUE}_{\text{min-max}}$ & $\mathbf{0.72 \pm 0.05}$ &	$\mathbf{0.77 \pm 0.07}$	& $\mathbf{0.73 \pm 0.05}$ \\
        \midrule
        \multicolumn{4}{c}{MS MARCO} \\
        \midrule
        Metric & Mistral-7B & Llama-2-7B & Llama-2-13B \\
        \midrule
        SinkRate (min)             & $0.73 \pm 0.03$ &	$0.70 \pm 0.02$ &	$0.72 \pm 0.04$ \\
        MaxEntropy               & $0.67 \pm 0.03$ &	$0.65 \pm 0.06$ &	$0.62 \pm 0.05$ \\
        $\text{INTRYGUE}_{\text{min-max}}$  & $\mathbf{0.77 \pm 0.03}$ & $\mathbf{0.72 \pm 0.04}$ &	$\mathbf{0.73 \pm 0.04}$ \\
        \bottomrule
    \end{tabular}%
    }
\end{table*}

\subsection{Sink Tokens}

\begin{figure}[tbp]
    \centering
    \begin{minipage}{0.48\textwidth}
        \centering
        \includegraphics[width=\linewidth]{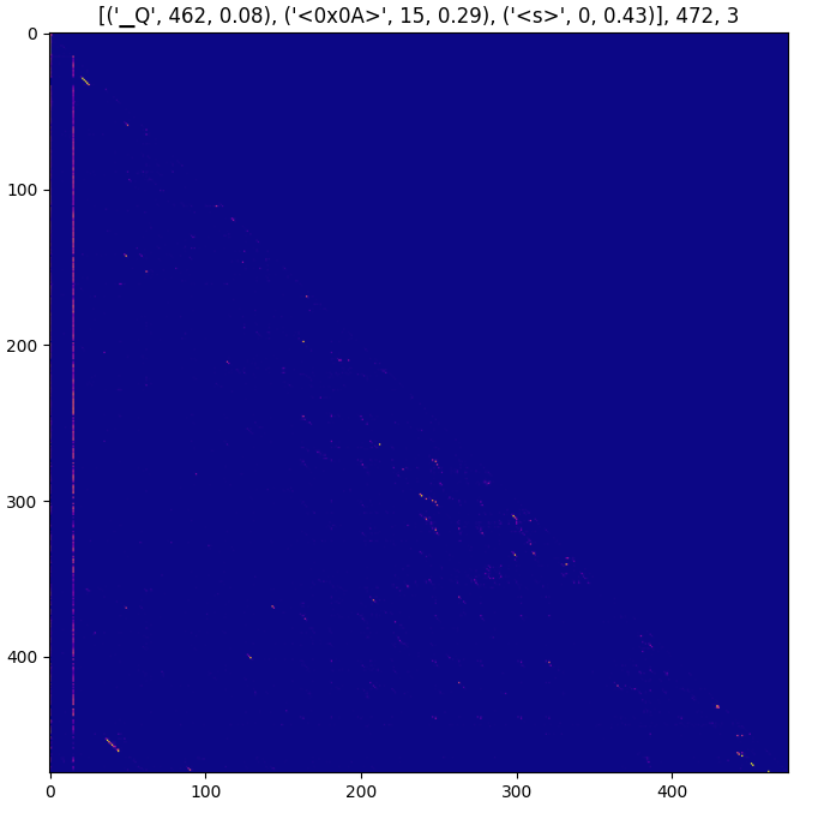}
        \caption{Attention matrix example with the sink patterns. Model: Mistral-7B, dataset: CoQA, head: L18H2.}
        \label{fig:subfig_a}
    \end{minipage}\hfill
    \begin{minipage}{0.48\textwidth}
        \centering
        \includegraphics[width=\linewidth]{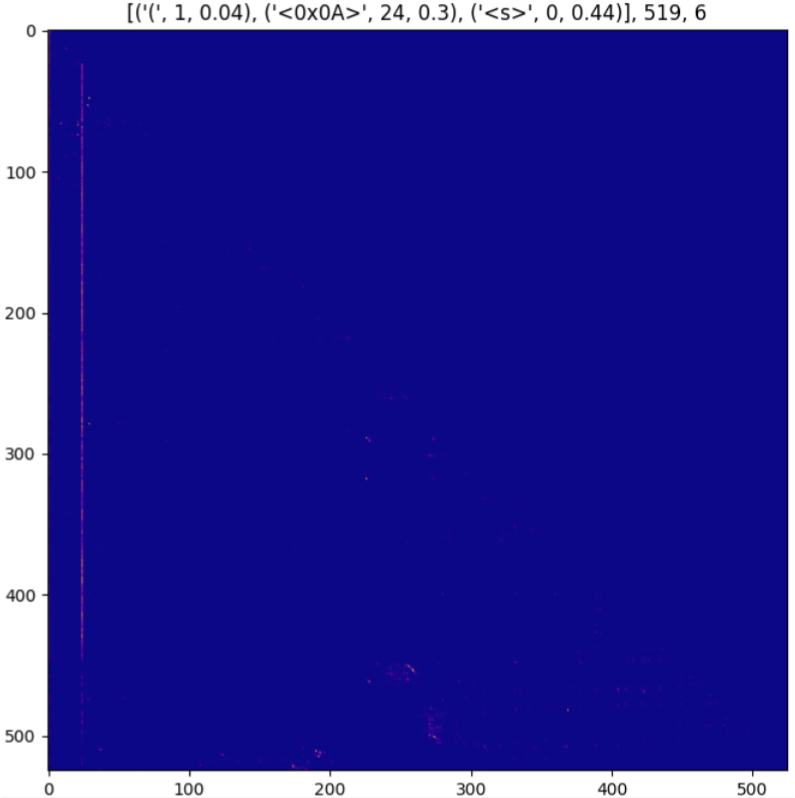}
        \caption{Attention matrix example with the sink patterns. Model: Mistral-7B, dataset: CoQA, head: L18H3.}
        \label{fig:subfig_b}
    \end{minipage}
    \label{fig:sink_tokens}
\end{figure}

Figures~\ref{fig:subfig_a} and \ref{fig:subfig_b} illustrate the sink patterns observed in Mistral-7B. The title of each plot lists the top three tokens with the highest SinkRate scores, formatted as (token, position, sink rate). Notably, attention sinks not only to the initial token but also to other utility tokens (e.g., \texttt{<0x0A>} in these examples).

\section{Implementation details}\label{app:implementation_details}

In this section, we describe the key implementation choices:

\begin{itemize}

\item For EigenScore, we used the representation of the last token to embed sentences, as suggested by~\cite{chen2024inside}. We extracted outputs from the 16th layer, as middle layers have been shown to contain the most factual information~\cite{sky2024androids, azaria-mitchell-2023-internal}.

\item All methods based on multiple generations employed $5$ additional model generations, with the exception of Semantic Density, which employed $10$, as a lower number of samples failed to achieve better-than-random prediction quality.

\item For a fair comparison with other baselines and our INTRYGUE score, the hyperparameters for the ReDeEP method were selected using response-wise labels (instead of token-wise labels, as proposed in the original paper).

\end{itemize}

All experiments were carried out across 5 random splits, corresponding to seeds $[42, \dots, 46]$. The train/val/test split fractions were set to 0.4/0.4/0.2, respectively. We used NVidia L40 and NVidia H100 GPUs.

\section{Use of scientific artifacts}

The CoQA dataset contains passages from seven domains operating under various licenses: literature and Wikipedia (CC BY-SA 4.0), MCTest children's stories (MSR-LA), RACE exam passages (custom license), and DeepMind CNN news passages (Apache). All dataset artifacts were utilized strictly following their intended licensing terms. The considered datasets contain no personally identifiable information or offensive content. Finally, the text of this paper was proofread and refined for clarity using Gemini-3.1-Pro.

\section{Potential risks}

\paragraph{Faithfulness vs. Factual Truth.} INTRYGUE measures how well an output is grounded in the retrieved text, not its objective truth. If a RAG pipeline retrieves false, biased, or malicious context, the model may faithfully reproduce it. INTRYGUE will accurately report low uncertainty in these cases, which end-users could dangerously misinterpret as a guarantee of factual correctness.

\paragraph{Adversarial Vulnerabilities.} Tying uncertainty scores to a specific mechanistic proxy introduces a theoretical attack vector. Adversaries could craft prompt injections specifically designed to artificially spike induction head activity, potentially tricking the gating mechanism into heavily suppressing the uncertainty score for manipulated outputs.

\end{document}